\documentclass{article}

\PassOptionsToPackage{numbers, compress}{natbib}
\usepackage[preprint]{neurips_2026}

\usepackage[utf8]{inputenc} 
\usepackage[T1]{fontenc}    
\usepackage{booktabs}       
\usepackage{amsfonts}       
\usepackage{nicefrac}       
\usepackage{microtype}      
\usepackage{xcolor}         

\usepackage{booktabs, amsmath, amsthm}
\usepackage{graphicx}
\usepackage{tikz-cd}
\usepackage{multirow}
\usepackage{enumitem}
\usepackage{subcaption}
\usepackage{hyperref}
\usepackage{url}            
\usepackage[capitalize,noabbrev]{cleveref}

\definecolor{mydarkblue}{rgb}{0,0.08,0.45}
\hypersetup{
  colorlinks=true,
  linkcolor=mydarkblue,
  citecolor=mydarkblue,
  urlcolor=mydarkblue
}

\usepackage{tikz}
\usetikzlibrary{calc}
\definecolor{pastelRed}{HTML}{FF6961}        
\definecolor{pastelPeachPink}{HTML}{fc8d62}   
\definecolor{pastelBlue}{HTML}{84B6F4}       
\definecolor{curveGreenDark}{HTML}{35B779}  
\definecolor{curveGreenLight}{HTML}{B0F2C2} 
\definecolor{darkGrayText}{gray}{0.3}

\newcommand{\drawsquare}[3]{%
    \fill[#2] ($(#1) + (-#3,-#3)$) rectangle ($(#1) + (#3,#3)$);
}

\title{On the Construction and Implications of\\ Low-Loss Valleys in LoRA-based Bayesian Inference}

\author{%
  Daniel Dold$^1$, Emanuel Sommer$^{2,3}$, Julius Kobialka$^{2,3}$, Oliver Dürr$^{1,4}$, David Rügamer$^{2,3}$ \\
  \normalfont
  $^1$ HTWG Konstanz $^2$ LMU Munich $^3$ Munich Center for Machine Learning (MCML) $^4$ TIDIT.ch \\
  \texttt{\{ddold, oliver.duerr\}@htwg-konstanz.de} \\ 
  \texttt{\{emanuel.sommer, julius.kobialka, david.ruegamer\}@stat.uni-muenchen.de}
}

\newcommand{\nseg}{N_{\text{seg}}}
\newcommand{\segdeg}{(m+1)}

\newcommand{\dout}{d_{\text{out}}}
\newcommand{\din}{d_{\text{in}}}
\newcommand{\lorarank}{r}
\newcommand{\deltaw}{\Delta W}

\newcommand{\curve}{\zeta} 
\newcommand{\noise}{\epsilon}

\newtheorem{proposition}{Proposition}

\begin{document}

\maketitle

\begin{abstract}
    While parameter-efficient fine-tuning methods like low-rank adaptation (LoRA) are standard for large language models, principled estimation of epistemic uncertainty remains challenging. Recent results in the LoRA regime suggest that discrete multi-mode approaches such as deep ensembles offer little benefit over single-mode methods. This contradicts broader observations in deep learning, where ensembling independent optima typically improves generalization, and linking these modes through continuous low-loss valleys further enhances Bayesian model averaging (BMA). Whether such structure exists in the LoRA space and whether it yields functional diversity missed by local or discrete methods has not been studied. We introduce \emph{LoRA-Curve}, a segmented B\'ezier curve parameterization in the LoRA space, with two variants: a free configuration that jointly optimizes all control points, and an anchored configuration that connects independently fine-tuned LoRA optima. 
    We prove pathwise continuity and Lipschitz regularity of the loss along the curve and empirically show, across reasoning and classification benchmarks with Qwen2.5\,7B, that linear interpolation encounters loss barriers, while our anchored multi-segment curves connect independent optima through continuous low-loss valleys. Combined with flat-minima perturbations and a Jensen--Shannon divergence regularizer, \emph{LoRA-Curve} yields measurably higher mutual information of the predictive distribution without sacrificing performance, and links continuous parameter-space traversal to functional diversity.
\end{abstract}

\section{Introduction}
Large language models (LLMs) have transformed language-based applications, yet they provide no principled estimate of epistemic uncertainty.  This becomes especially problematic during fine-tuning,  where point estimates lead to overconfident predictions and suboptimal generalization \citep[see, e.g.,][]{li2025calibrating}. 
To address this, probabilistic methods such as variational inference (VI) have been proposed; in the setting of low-rank adaptation (LoRA) \citep{huLoRA2022}, capturing a single, well-behaved mode has proven highly effective \citep{samplawskiScalable2025,wangBLoB2024}.
The reported results in \citet{wangBLoB2024, samplawskiScalable2025} surprisingly show superior performance compared to deep ensembles \citep{lakshminarayananSimple2017},
which contradicts broader findings where functional diversity stems from multiple distinct optima rather than a single basin. 
Existing Bayesian LoRA methods represent two extremes: either a single local mode (VI) or a discrete collection of isolated modes (DE). These optima can be connected by non-linear, low-loss paths \citep{garipovLoss2018, draxler2018essentially}, forming a natural subspace for Bayesian model averaging (BMA) \citep{doldPaths2025,izmailovSubspace2020}. 
%

Whether a continuous low-loss structure links these LoRA optima, and whether marginalizing over it provides functional diversity beyond these extremes, has not been studied in the context of LoRA.
We therefore construct low-loss valleys in the LoRA weight space via B\'ezier curves initially used for mode connectivity \citep{garipovLoss2018}, which simultaneously enable principled BMA. 

This allows us to address the following research questions: 1) \emph{Is it possible to construct meaningful low-loss valleys in the LoRA weight space?} 2) \emph{Does continuous mode connectivity capture functional diversity missed by both local variational methods and discrete ensembles?} And 3) \emph{What do these findings imply for probabilistic LoRA approaches?}
\begin{figure}[t]
  \centering
  \resizebox{\linewidth}{!}{%
  \begin{tikzpicture}[x=1cm,y=1cm,line cap=round,line join=round]

    \begin{scope}[shift={(0.0, 0)}]
        
        \draw[rounded corners=0.3cm, fill=gray!5, draw=gray!30, thick] (0.6, 0.8) rectangle (6.8, 4.3);
        \node[gray!50, anchor=north east] at (6.8, 4.3) {\small query$_{L+1}$};

        \draw[rounded corners=0.3cm, fill=gray!10, draw=gray!40, thick] (0.3, 0.4) rectangle (6.5, 3.9);
        \node[gray!60, anchor=north east] at (6.5, 3.95) {\small value$_{L}$};

        \draw[rounded corners=0.3cm, fill=white, draw=gray!70, thick] (0, 0) rectangle (6.2, 3.5);
        \node[gray!80, anchor=north east] at (6.2, 3.5) {\small query$_L$};

        \begin{scope}[shift={(0, 0)}]
            
            \draw[thick, draw=gray!70, fill=white, rounded corners=0.1cm] (0.4, 0.5) rectangle (0.8, 2.5);
            \node[darkGrayText, font=\normalsize] at (0.6, 1.5) {$X$};

            \draw[thick, draw=gray!70, fill=gray!5, rounded corners=0.1cm] (1.3, 0.3) rectangle (4.6, 1.3);
            \node[darkGrayText, font=\normalsize] at (2.95, 0.8) {$W_0^T \in \mathbb{R}^{d_{\mathrm{in}} \times d_{\mathrm{out}}}$};

            \draw[thick, draw=pastelBlue, fill=pastelBlue!15, shift={(0.3, 0.3)}]
                (1.3, 2.5) -- (2.3, 2.3) -- (2.3, 1.7) -- (1.3, 1.5) -- cycle;
            \draw[thick, draw=pastelRed, fill=pastelRed!15, shift={(0.15, 0.15)}]
                (1.3, 2.5) -- (2.3, 2.3) -- (2.3, 1.7) -- (1.3, 1.5) -- cycle;
            \draw[thick, draw=pastelPeachPink, fill=pastelPeachPink!15, shift={(0, 0)}]
                (1.3, 2.5) -- (2.3, 2.3) -- (2.3, 1.7) -- (1.3, 1.5) -- cycle;
                
            \node[darkGrayText, font=\normalsize, anchor=east] at (2., 2.0) {$\mathbf{A}$};

            \draw[thick, draw=pastelBlue, fill=pastelBlue!15, shift={(0.3, 0.3)}]
                (3.3, 2.3) -- (4.3, 2.5) -- (4.3, 1.5) -- (3.3, 1.7) -- cycle;
            \draw[thick, draw=pastelRed, fill=pastelRed!15, shift={(0.15, 0.15)}]
                (3.3, 2.3) -- (4.3, 2.5) -- (4.3, 1.5) -- (3.3, 1.7) -- cycle;
            \draw[thick, draw=pastelPeachPink, fill=pastelPeachPink!15, shift={(0, 0)}]
                (3.3, 2.3) -- (4.3, 2.5) -- (4.3, 1.5) -- (3.3, 1.7) -- cycle;
                
            \node[darkGrayText, font=\normalsize, anchor=west] at (3.6, 2.0) {$\mathbf{B}$};

            \node[circle, draw=gray!70, thick, fill=white, inner sep=1.pt] (plus) at (5.0, 1.5) {\small $+$};

            \draw[thick, draw=gray!70, fill=white, rounded corners=0.1cm] (5.4, 0.5) rectangle (5.8, 2.5);
            \node[darkGrayText, font=\normalsize] at (5.6, 1.5) {$Y$};

            \coordinate (Xin) at (0.8, 1.5);
            \coordinate (Ain) at (1.3, 2.0);
            \coordinate (Aout) at (2.3, 2.0);
            \coordinate (Bin) at (3.3, 2.0);
            \coordinate (Bout) at (4.6, 2.3);
            \coordinate (W0in) at (1.3, 0.8);
            \coordinate (W0out) at (4.6, 0.8);
            \coordinate (Yin) at (5.4, 1.5);

            \draw[->, >=stealth, thick, gray!70] (Xin) -- ++(0.2,0) |- (Ain);
            \draw[->, >=stealth, thick, gray!70] (Xin) -- ++(0.2,0) |- (W0in);
            \foreach \i/\arrow in {0/->, 0.15/-, 0.3/-} {
                \draw[\arrow, >=stealth, thick, gray!70, dashed]
                    ([shift={(\i,\i)}]Aout) --
                    ([shift={(0,\i)}]Bin);
            }
            \draw[->, >=stealth, thick, gray!70] (Bout) -| (plus.north);
            \draw[->, >=stealth, thick, gray!70] (W0out) -| (plus.south);
            \draw[->, >=stealth, thick, gray!70] (plus.east) -- (Yin);
            
        \end{scope}

        \node[darkGrayText, font=\normalsize] at (2.95, 3.1) {$\{A, B\} \subset \theta \in \Theta$};
    \end{scope}

    \begin{scope}[shift={(7.0, 0.0)}] 
        \def\XBoxLength{2.2}

        \begin{scope}[shift={(0, 3.0)}]
            \draw[rounded corners=0.25cm, gray!70, line width=0.3pt] (0, 0) rectangle (\XBoxLength, 1.3);
            \node[darkGrayText, font=\small, anchor=north west] at (0.1, 1.3) {MAP};
            
            \fill[pastelPeachPink] (1.2, 0.9) circle (0.08);
            \node[darkGrayText, font=\scriptsize, align=center] at (1.2, 0.35) {$p(\theta\mid\mathcal D) \approx$ \\ $\delta(\theta-\hat\theta_{\mathrm{map}})$};
        \end{scope}

        \begin{scope}[shift={(0, 1.5)}]
            \draw[rounded corners=0.25cm, gray!70, line width=0.3pt] (0, 0) rectangle (\XBoxLength, 1.4);
            \node[darkGrayText, font=\small, anchor=north west] at (0.1, 1.4) {MFVI};
            
            \draw[pastelPeachPink, line width=0.8pt] (1.2, 0.7) ellipse (0.6 and 0.25);
            \draw[pastelPeachPink, line width=0.8pt] (1.2, 0.7) ellipse (0.3 and 0.125);
            \fill[pastelPeachPink] (1.2, 0.7) circle (0.05);
            \node[darkGrayText, font=\scriptsize, align=center] at (1.2, 0.25) {$p(\theta\mid\mathcal D) \approx q(\theta)$};
        \end{scope}

        \begin{scope}[shift={(0, 0)}]
            \draw[rounded corners=0.25cm, gray!70, line width=0.3pt] (0, 0) rectangle (\XBoxLength, 1.4); 
            \node[darkGrayText, font=\small, anchor=north west] at (0.1, 1.4) {DE};
            
            \fill[pastelPeachPink] (0.6, 0.8) circle (0.08); 
            \fill[pastelRed] (1.2, 1.0) circle (0.08);       
            \fill[pastelBlue] (1.8, 0.8) circle (0.08);      
            \node[darkGrayText, font=\scriptsize, align=center] at (1.2, 0.35) {$p(\theta\mid\mathcal D) \propto$ \\ $\sum_{i=1}^3 \delta(\theta-\theta_i)$};
        \end{scope}

    \end{scope}

    \begin{scope}[shift={(9.4, -0.3)}] 
        \begin{scope}
            \def\curveboxYmin{-0.1}
            \def\curveboxYmax{2.4}
            \def\XBoxLength{5.7}
            
            \coordinate (modeAint) at (0.45, .9);      
            \coordinate (modeBint) at (2.85, 1.3);      
            \coordinate (modeCint) at (5.3, 0.9);      
            \coordinate (cpAint) at   (1.6, 1.8);        
            \coordinate (cpBint) at   (4.6, 2.);        
            \coordinate (cpAcurve) at (1.62, 1.62);        
            \coordinate (cpBcurve) at (4.51, 1.75);        

            \draw[curveGreenDark, densely dashed, thick] (0.45, 3.2) -- (modeAint);
            \draw[curveGreenDark, densely dashed, thick] (1.6, 3.2) -- (cpAcurve);
            \draw[curveGreenDark, densely dashed, thick] (2.85, 3.2) -- (modeBint);
            \draw[curveGreenDark, densely dashed, thick] (4.1, 3.2) -- (cpBcurve);
            \draw[curveGreenDark, densely dashed, thick] (5.3, 3.2) -- (modeCint);

            \def\tBoxOffset{0.2}
            \def\tBoxYLength{0.4}
            \draw[gray, fill=white, thick, rounded corners=2pt] (.2, \curveboxYmax + \tBoxOffset) -- (\XBoxLength - 0.2, \curveboxYmax + \tBoxOffset) -- (\XBoxLength - 0.2, \curveboxYmax + \tBoxOffset + \tBoxYLength) -- (.2, \curveboxYmax + \tBoxOffset + \tBoxYLength) -- cycle;
            \node[darkGrayText, font=\normalsize] at (2.3, \curveboxYmax + \tBoxOffset + \tBoxYLength/2 ) {$t$};
            \draw[gray, thick, -, >=stealth] (2., \curveboxYmax + \tBoxOffset + \tBoxYLength/2) circle (0.15); 
            \draw[gray, thick, -, >=stealth] (2., \curveboxYmax + \tBoxOffset + \tBoxYLength/2) -- (1.9, \curveboxYmax + 0.1 + \tBoxOffset);
            \draw[<->, thick, gray] (2.9, \curveboxYmax + \tBoxOffset + \tBoxYLength/2) -- (3.7, \curveboxYmax + \tBoxOffset + \tBoxYLength/2);
            \drawsquare{3.3, \curveboxYmax + \tBoxOffset + \tBoxYLength/2}{curveGreenDark}{0.08};

            \node[anchor=south west, inner sep=0pt] (hm) at (-0.15, 3.1) {\includegraphics[width=\XBoxLength cm]{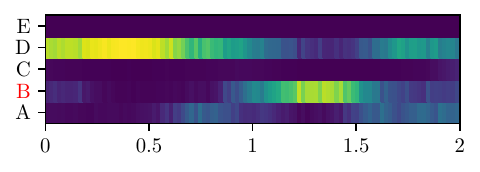}};
                        
            \foreach \x in {0.4, 1.62, 2.85, 4.1, 5.3} {
                \draw[curveGreenDark, thick, rounded corners=2pt] 
                    (\x-0.1, 3.55) rectangle (\x+0.1, 5.1);
            }

            \draw[rounded corners=0.25cm, gray, line width=.6pt] (0.0, \curveboxYmin) rectangle (\XBoxLength, \curveboxYmax);
            \node[darkGrayText, font=\normalsize, anchor=north west, align=left, fill=white] at (0.12, \curveboxYmax-0.04) {Curve {\em ours}};

            \draw[gray!60, dashed, thin] (modeAint) -- (cpAint) -- (modeBint);
            \draw[gray!60, dashed, thin] (modeBint) -- (cpBint) -- (modeCint);
            
            \draw[curveGreenDark, line width=1.0pt] (modeAint) .. controls (cpAint) .. (modeBint);
            \draw[curveGreenDark, line width=1.0pt] (modeBint) .. controls (cpBint) .. (modeCint);
                        
            \fill[pastelPeachPink] (modeAint) circle (0.16); 
            \node[text=pastelPeachPink, xshift=3.5mm, yshift=-0.1mm, font=\footnotesize] at (modeAint) {$\theta_1$};
            \fill[pastelRed] (modeBint) circle (0.16);              
            \node[text=pastelRed, xshift=0mm, yshift=-3.2mm, font=\footnotesize] at (modeBint) {$\theta_3$};
            \fill[pastelBlue] (modeCint) circle (0.16);      
            \node[text=pastelBlue, xshift=-3.2mm, yshift=0mm, font=\footnotesize] at (modeCint) {$\theta_5$};
            \fill[curveGreenLight] (cpAint) circle (0.1); 
            \fill[curveGreenLight] (cpBint) circle (0.1); 

            \drawsquare{modeAint}{curveGreenDark}{0.08}
            \drawsquare{modeBint}{curveGreenDark}{0.08}
            \drawsquare{modeCint}{curveGreenDark}{0.08}
            \drawsquare{cpAcurve}{curveGreenDark}{0.08}
            \drawsquare{cpBcurve}{curveGreenDark}{0.08}

            \node[darkGrayText, font=\footnotesize, align=center] (seg1) at (1.875, 1.2) {segment 1};
            \draw[darkGrayText, thin, shorten >=1pt] (seg1.north) -- (1.875, 1.64);

            \node[darkGrayText, font=\footnotesize, align=center] (seg2) at (4., 1.2) {segment 2};
            \draw[darkGrayText, thin, shorten >=1pt] (seg2.north) -- (4., 1.77);

            \node[darkGrayText, font=\small] at (\XBoxLength/2, .55) {
                \textcolor{pastelPeachPink}{$N=3$} anchors; \textcolor{curveGreenDark}{$m=1$} handles / seg
            };
            \node[darkGrayText, font=\footnotesize] at (\XBoxLength/2, 0.1) {$p(\theta\mid\mathcal D) \propto \int \delta(\theta-\theta(t))dt$};
        \end{scope} 
    \end{scope}

  \end{tikzpicture}
  }
    \caption{Overview of our LoRA-Curve approach. \textbf{Left:} Architecture for linear multiplication in an attention layer using low-rank adapters $A$ and $B$, alongside frozen base weights $W_0^T$ for different layers $L$. \textbf{Middle:} Comparison of Bayesian approximation strategies. Maximum A Posteriori (MAP) and mean-field variational inference (MFVI) capture only a single mode, whereas deep ensembles (DE) capture multiple distinct modes (colored orange, red, and blue). \textbf{Right:} Our proposed continuous parameterization. The framework connects $N=3$ anchor points (e.g., DE modes) using a two-segment curve shaped by intermediate handles (green, $m=1$ per segment), where anchors can either be fixed or jointly optimized with handles (bottom right). Mapping the curve parameter $t$ reveals smooth, continuous transitions in predictive class probabilities between the discrete modes (top right). The green squares visualize the correspondence between the curve position and the resulting probabilities.}
  \label{fig:figure_1}
\end{figure}
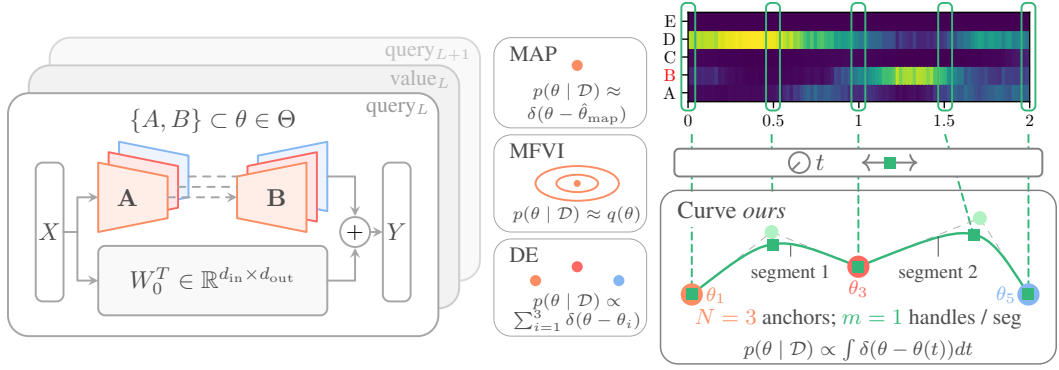
\paragraph{Contributions} We thereby make the following contributions:
\begin{itemize}[leftmargin=1.5em]
\item We introduce \emph{LoRA-Curve}, a framework for constructing continuous, segmented paths that explicitly enable smooth, functionally diverse traversals through the LoRA space.
\item We demonstrate that continuous mode connectivity captures {cross-modal functional diversity} that discrete ensembles only partially exploit
, while attaining competitive generalization performance.
\item We characterize the resulting loss valleys and relate these findings to complementary approaches to understand the key properties of the LoRA posterior space.
\end{itemize}

\section{Background and related work}

Fine-tuning LLMs is highly resource-intensive, prompting the widespread adoption of parameter-efficient fine-tuning methods like LoRA \citep{huLoRA2022}. 
Given a layer with weights $W\in\mathbb{R}^{\dout \times \din}$, LoRA \citep{huLoRA2022} can be used to adapt a pre-trained weight $W_0\in\mathbb{R}^{\dout\times \din}$ by defining fine-tuning parameters $A \in\mathbb{R}^{\lorarank\times \din}$ and $B \in\mathbb{R}^{\dout\times \lorarank}$ with dimension $\lorarank \ll \min(\dout, \din)$ and learning the update
\begin{equation} \label{eq:lora}    
W = W_0 + \deltaw, \qquad \deltaw = BA.
\end{equation}
During the fine-tuning phase, only $A$ and $B$ matrices are optimized and $W_0$ is kept fixed. While this approach significantly reduces the number of trainable parameters, standard LoRA fine-tuning can still converge to sharp minima that generalize poorly \citep{liFlatLoRA2025}. To explicitly encourage the optimization process to settle in wide, generalizable regions of the loss landscape, \citet{liFlatLoRA2025} suggested a perturbation strategy called Flat-LoRA, which extends \cref{eq:lora} by injecting noise $\noise$ during training. The modified update rule then becomes $W = W_0 + \deltaw + \noise$ where $\noise \in \mathbb{R}^{\dout \times \din}$ with entries $\noise_{i,j} \sim \mathcal{N}(0,\frac{\rho}{\din} \|W'_{i,:}\|^2_2)$, $W'_{i,:}$ the $i$th row of the $W'=W_0+\Delta W$, and hyperparameter $\rho>0$.
In the following, we will denote all (flattened) LoRA parameters, potentially gathered across multiple layers, with $\theta\in\Theta\subseteq \mathbb{R}^D$ with $D$ indicating the number of trainable LoRA parameters.

\paragraph{Bayesian parameter-efficient fine-tuning}

Classic approaches for capturing predictive uncertainty in deep learning rely on variational inference (VI) \citep[see, e.g.,][]{blundellWeight2015,ranganathBlack2014}, the Laplace approximation \citep[e.g.,][]{daxbergerLaplace2021}, and deep ensembles (DEs) \citep{lakshminarayananSimple2017}. 
To scale these techniques to modern LLMs, recent work has integrated uncertainty estimation in the {LoRA} parameterization. Prominent variational examples include \emph{Bayesian Low-Rank Adaptation by Backpropagation} ({BLoB}) \citep{wangBLoB2024}, which learns a variational distribution over the $A$ matrix via backpropagation, and \emph{Scalable Bayesian Low-Rank Adaptation} ({ScalaBL}) \citep{samplawskiScalable2025}, which utilizes further weight factorization to drastically reduce the stochastic parameter count for improved scalability. Other approaches include discrete LoRA ensembles to approximate the posterior \citep{balabanovUncertainty2025,yangBayesian2024}.

\paragraph{Loss landscapes and mode connectivity}
While the previously mentioned approaches successfully capture epistemic uncertainty at a fraction of the full-parameter cost, their geometric assumptions are in stark contrast (single- vs.\ multi-mode) and conflict with each other regarding the geometry of the posterior or loss landscape. In general neural network applications, it has been shown that independent neural network optima are rarely isolated, but rather often form connected and continuous paths of low loss \citep{draxler2018essentially,garipovLoss2018}. The analysis of low-loss valleys could therefore provide insights into the geometrical structure of LoRA posterior spaces, but so far, subspace methods have not been (successfully) applied to LoRA. 

\section{Constructing a curve-subspace for LoRA-based inference}
We demonstrate how to efficiently construct continuous, low-loss valleys within the LoRA weight space. To this end, we introduce \emph{LoRA-Curve}, a segmented Bézier parameterization for the LoRA space. Depending on the initialization and optimization of the anchor control points of the curve, we will distinguish between two configurations: \textbf{Free LoRA-Curve}, where all control points are jointly optimized, and \textbf{Anchored LoRA-Curve}, where segment boundaries (anchor points) are frozen to independently pretrained adapters (see \cref{fig:figure_1}).

\subsection{Free LoRA-Curve}
\label{subsec:flc}
Building on the flexibility of different parameterizations discussed in \citet{garipovLoss2018}, we construct the LoRA-Curve using a B\'ezier curve with $m+2$ control points. Specifically for the LoRA fine-tuning setting, we parameterize a continuous path $t \mapsto \curve_{\vartheta}(t) \in \Theta$ for $t \in [0,1]$ using 
$$
\curve_{\vartheta}(t) = \textstyle\sum_{i=0}^{m+1} b_{i,\segdeg}(t) \theta_i ,$$ 
where $\vartheta := \{ \theta_{0}, \dots, \theta_{\segdeg}\}$ represents the set of all control points $\theta_i \in \Theta$, and $b_{i,\segdeg}(t) = \binom{\segdeg}{i} (1-t)^{\segdeg-i} t^i$ are the Bernstein basis polynomials for a curve of degree $\segdeg$. 
Because each $\theta$ has a one-to-one mapping to the low-rank matrices $A$ and $B$ within a given layer, $\curve_{\vartheta}(t)$ directly determines the continuous matrices $A_{\vartheta}(t)$ and $B_{\vartheta}(t)$.
Consequently, the updated weight matrix at any point along the curve is $W_{\vartheta}(t) = W_0 + B_{\vartheta}(t)A_{\vartheta}(t)$.
This curve defines an adaptation trajectory in the LoRA space: only $A$ and $B$ vary along $t$, while the pretrained weights $W_0$ remain fixed. Since all control points are jointly trained from scratch, we refer to this as the \emph{Free LoRA-Curve} (FLC).

\paragraph{Optimization}
We train the curve by optimizing all control points $\vartheta$ using the expected loss along the curve
\begin{equation}
    \mathcal{L}_{\text{curve}}\left( \vartheta \right) := \mathbb{E}_{t \sim U(0,1)} [ \mathcal{L}_{\text{CE}}(\mathcal{D}; W_{\vartheta}(t)) ],
    \label{eq:objective}
\end{equation}
where $\mathcal{L}_{\text{CE}}$ is the negative categorical log-likelihood averaged over the training data $\mathcal{D}$. 
Unlike the standard continuous curve formulation of \citet{garipovLoss2018}, which optimizes a path between fixed, pre-trained endpoints, we jointly optimize the entire curve across all control points $\vartheta$. To maintain standard LoRA behavior at initialization, we configure each control point $\theta_i$ by drawing the respective LoRA matrix $A_i$ randomly and setting $B_i$ to zero. This ensures $B_i A_i = 0$ for all $i$, so the initial weights evaluate exactly to $W_0$ everywhere along the curve.

During training, we approximate the expectation in \cref{eq:objective} by sampling a new $t \sim U(0,1)$ for every minibatch. Compared to an $m+2$-ensemble approach, our approach maintains an identical memory footprint but is less expensive to train, as it can update all curve members simultaneously and only requires a linear combination of control points to compute $\curve_{\vartheta}(t)$, followed by a single forward pass per $t$-sample. Finally, to ensure the optimized curve settles in a flat low-loss valley, we incorporate the method of \citet{liFlatLoRA2025} by injecting perturbation noise $\noise$ into $W$ with $W=W_{\vartheta}(t) + \noise$.

\paragraph{Inference}
During inference for a new data point $(x^\ast,y^\ast)$, we approximate the posterior predictive distribution via a temperature-scaled grid approximation along the optimized curve:
$$
    p(y^\ast \mid x^\ast, \mathcal{D}) \approx \sum_{j=1}^{M} \varpi_j(T)\, p\left(y^\ast \mid x^\ast, W_{\vartheta}(t_j), \mathcal{D}\right), 
    \quad \varpi_j(T) = \frac{p(\mathcal{D} \mid t_j)^{1/T}}{\sum_{k=1}^{M} p(\mathcal{D} \mid t_k)^{1/T}},
$$
where $\{t_j\}_{j=1}^M$ are $M$ equispaced grid points. We empirically find that $T \to \infty$ yields the most robust performance across datasets. This corresponds to uniform averaging $\varpi_j = 1/M$. This aligns with the intuition that the curve has already been optimized on $\mathcal{D}$, so reweighting by a posterior over $t$ would effectively condition on the training data a second time, thereby collapsing the predictive diversity induced by the curve. We provide further details in Appendix~\ref{sec:appendix_temperature}.

\subsection{Segmented curves}
\label{subsec:seg_curve}
To navigate highly non-convex loss landscapes, a single polynomial segment is often insufficiently flexible (cf.\ \cref{fig:plot1}). We therefore extend the LoRA-Curve framework to a composite B\'ezier curve comprising multiple individual segments. 
To construct this composite path, we define two distinct types of control points: \emph{anchors}, which act as segment boundaries and lie directly on the curve, and \emph{handles}, which are intermediate points that shape the curve but generally do not lie on it (cf.\ \cref{fig:figure_1}).

Let $N$ denote the number of anchors and $m$ the number of handles per segment.
For a curve with $N$ anchor points connected by $\nseg =N-1$ segments, each with $m$ handles, the global parameter is $t \in [0, \nseg]$ and the total number of control points is $N_{\text{cp}} = \nseg (m+1) + 1$. The active segment index is determined via $k = \min(\lfloor t \rfloor, N-2) \in \{0, \dots, N-2\}$ with local coordinate $\tau = t - k \in [0, 1]$. We then evaluate the curve locally using this subset of control points
$
\curve_{\vartheta}(\tau) = \sum_{i=0}^{m+1} b_{i,m+1}(\tau)\, \theta_{k(m+1) + i}.
$
This piecewise construction ensures that adjacent segments share exactly one boundary control point (an anchor point). This construction guarantees $C^0$ continuity while permitting sharp corners, allowing the path to flexibly bend around steep loss barriers (cf.\ the bend at $\theta_3$ in \cref{fig:figure_1}). By contrast, a single B\'ezier curve of degree $(N_{\text{cp}}-1)$ is globally coupled: every control point influences the entire curve via its Bernstein basis function, making localized bends difficult and the path overly rigid when navigating highly non-convex valleys.


\subsection{Anchored LoRA-Curve}
\label{subsec:alc}
While FLC optimizes all control points (the $N$ anchor points and all $\nseg m$ handles) from scratch, we can explicitly guide the curve to connect known high-performing optima (e.g., those obtained from a Deep Ensemble) by freezing the anchors to the weights of independently fine-tuned LoRA models. 
In this scenario, the remaining handles (initialized exactly as in FLC to explicitly increase diversity) are updated via gradient descent according to \cref{eq:objective}. 
We call this approach \emph{Anchored LoRA-curve (ALC)} and its configuration as $\text{ALC}(N, m)$, where $N$ is the number of frozen anchors and $m$ the number of handles per segment. Analogously, $\text{FLC}(N, m)$ denotes the free variant where the anchors and handles are jointly optimized.

This unified notation mathematically subsumes several standard methods as special cases: $\text{MAP}$ corresponds to $\text{FLC}(1, 0)$ (a single point estimate with no curve); $\text{DE}(N)$ corresponds to $N$ independent anchors with $m=0$; and $\text{ALC}(N, 0)$ reduces to standard piecewise linear mode connectivity between $N$ optima, which we denote as $\text{Lin}(N)$.

\subsection{Diversity regularization}
\label{subsec:jsd_math}
To encourage functional diversity along the path and prevent the curve from collapsing into a single point-estimate, we employ a margin-based Jensen-Shannon Divergence (JSD) regularizer. Let $f(x; W_{\vartheta}(t))$ denote the model logits for input $x$ at curve parameter $t \in [0, \nseg]$, and let $p_t := \operatorname{softmax}(f(x; W_{\vartheta}(t)))$ denote the resulting predictive distribution. 
During each training step, we sample a point $t_1 \sim U(0, \nseg)$ and compute an offset $t_2 = (t_1 + \nseg/2) \pmod{\nseg}$. This ensures that $t_1$ and $t_2$ are well separated along the curve's parameterization. We then compute the predictive distributions $p_{t_1}$ and $p_{t_2}$ and their mixture $\overline{p} = \frac{1}{2}(p_{t_1} + p_{t_2})$. The JSD between the two predictive distributions is
$
    D_{\mathrm{JSD}}(p_{t_1} \parallel p_{t_2}) = \frac{1}{2} ( D_{\mathrm{KL}}(p_{t_1} \parallel \overline{p}) + D_{\mathrm{KL}}(p_{t_2} \parallel \overline{p}) ).
$ 
To prevent unbounded JSD regularization from degrading primary task performance, we clip the penalty to $\tau_{\mathrm{JSD}}$.
The total loss optimized at each step is then the symmetric cross-entropy loss augmented by this diversity penalty:
\begin{equation*}
    \mathcal{L} = \frac{1}{2} \Big( \mathcal{L}_{\mathrm{CE}}(t_1) + \mathcal{L}_{\mathrm{CE}}(t_2) \Big) + \lambda_{\mathrm{JSD}} \mathcal{L}_{\mathrm{JSD}},
    \label{eq:total_loss}
\end{equation*}
where $\lambda_{\mathrm{JSD}}$ controls the regularization strength. By explicitly penalizing the similarity of predictions at distant points on the curve, the JSD regularizer forces the trajectory to explore diverse functional regions of the loss valley.

\section{Properties of Bayesian inference using LoRA-Curve}
\label{sec:lora_curve_properties}

While connecting well-performing solutions might improve the quality of performance and uncertainty quantification, we see the primary merit of the curve in its properties, which allow us to connect the parameter to the output space and interpret the model's posterior geometry.

\subsection{Continuity}

In contrast to DEs or samples from a variational distribution that only provide access to a finite set of isolated predictive distributions
$\{p(y \mid x, \theta_i)\}_{i=1}^{N}$, the construction of a curved subspace allows users to understand how predicted probabilities update when making changes in the LoRA weight space. A straightforward but important result is the following.

\begin{proposition}
\label{prop:continuity_main}
The output distribution of the LoRA-Curve $p(y | x, W_{\vartheta}(t), \mathcal{D})$ is continuous in $t$
\end{proposition}
This result follows as the composition
\begin{equation*}
    t \;\mapsto\; \zeta_{\vartheta}(t)
      \;\mapsto\; W_0 + B_{\vartheta}(t)\,A_{\vartheta}(t)
      \;\mapsto\; f\bigl(x; W_{\vartheta}(t)\bigr)
      \;\mapsto\; p(y | x, W_{\vartheta}(t), \mathcal{D})
\end{equation*}
is continuous in each step:
$\zeta_{\vartheta}(t) = \sum_{i=0}^{m+1} b_{i,m+1}(t)\, \theta_i$
is polynomial in $t$ (hence $C^{\infty}$);
the recovery of $A_{\vartheta}(t)$ and $B_{\vartheta}(t)$ from
$\zeta_{\vartheta}(t)$ is simply a projection onto fixed index
subsets of $\mathbb{R}^D$, and is therefore continuous;
the product $B_{\vartheta}(t)\,A_{\vartheta}(t)$ is bilinear and continuous in
both factors;
the forward pass $f(\,\cdot\,; W)$ is continuous in $W$ for activations such as SiLU used in LLMs;
and the softmax is continuous. Furthermore, for segmented curves, $C^0$ continuity at anchor points holds by
construction, since adjacent segments share a boundary control point
(cf.~Section~\ref{subsec:seg_curve}).

As depicted in \cref{fig:logits_smootheness}, this property provides direct access to the roles of weights in the LoRA space by relating the change in the low-loss valley to the change in output probabilities. The subspace thus not only provides a way to perform probabilistic inference in the LoRA space, but can also be used to interpret the importance and role of different weights.

\subsection{Posterior geometry along the curve}

While the successful application of ALC and FLC as demonstrated in the next section suggests the existence of a continuous low-loss valley, continuity alone does not quantify how predictions and losses change along the curve. We therefore study the empirical risk restricted to a fixed LoRA-Curve, $\ell_{\vartheta}(t)
    :=
    \mathcal{L}_{\mathrm{CE}}(\mathcal{D}; W_{\vartheta}(t)), t \in [0,\nseg]$.
Rather than requiring the restricted loss to be piecewise polynomial, which is not generally true for modern architectures involving softmax, SiLU, GELU, or normalization layers, we only require the regularity of the composed neural network map.

\begin{proposition}
\label{prop:pathwise_regular_curve}
Assume that the LoRA-Curve $t \mapsto W_{\vartheta}(t)$ is piecewise continuously differentiable on $[0,\nseg]$ and that, for every datapoint $(x,y)\in\mathcal{D}$, the map
 $W \mapsto \mathcal{L}_{\mathrm{CE}}(y, f(x;W))$
is continuously differentiable on an open neighborhood of the curve image $
    \{W_{\vartheta}(t): t\in[0,\nseg]\}$.
Then the restricted curve loss $\ell_{\vartheta}$ is continuous and piecewise continuously differentiable. On every differentiable segment, $
    \ell_{\vartheta}'(t)
    =
    \left\langle
        \nabla_W \mathcal{L}_{\mathrm{CE}}(\mathcal{D}; W_{\vartheta}(t)),
        \frac{d}{dt} W_{\vartheta}(t)
    \right\rangle$.
Moreover, since the curve image is compact, $\ell_{\vartheta}$ is Lipschitz on each segment and absolutely continuous on $[0,\nseg]$. In particular, for all $s,t$ in the same segment,
$$
    |\ell_{\vartheta}(t)-\ell_{\vartheta}(s)|
    \leq
    \int_s^t
    \left\|
        \nabla_W \mathcal{L}_{\mathrm{CE}}(\mathcal{D}; W_{\vartheta}(u))
    \right\|
    \|
        \frac{d}{du}W_{\vartheta}(u)
    \|
    du .
$$
\end{proposition}

This result follows directly from the chain rule and compactness. The LoRA-Curve is piecewise continuously differentiable because it is constructed from polynomial curve segments in the low-dimensional LoRA parameter space, followed by the affine-bilinear map $(A,B) \mapsto W_0 + BA$.
Modern LLM components such as affine maps, matrix products, and SiLU or softmax activations are continuous and smooth on their natural domains. Therefore, the loss restricted to the LoRA-Curve inherits this regularity.

The proposition provides a conservative but useful sense in which the LoRA-Curve exposes local posterior geometry. The loss cannot jump along the curve, and its change between two curve locations is controlled by two quantities: the norm of the full loss gradient and the speed with which the curve moves through weight space. Thus, sharp changes in prediction or loss along the curve must be explained either by large local sensitivity of the model, measured by
$
    \left\|
        \nabla_W \mathcal{L}_{\mathrm{CE}}(\mathcal{D}; W_{\vartheta}(t))
    \right\|,
$
or by rapid movement of the LoRA-Curve in parameter space, measured by the curve's speed
$
    \|
        \frac{d}{dt}W_{\vartheta}(t)
    \|.
$

From an inference perspective, this means that predictive distributions sampled along the curve are not isolated points, as in deep ensembles, but form a continuous, pathwise-regular trajectory. The curve, therefore, provides an interpretable one-dimensional probe of the local posterior geometry: it connects high-performing parameter settings while making changes in predictive probabilities attributable to movement in the LoRA space.

\section{Experimental study}

%
To answer our core research questions, our evaluation proceeds in two stages. First, we analyze the geometric properties of the LoRA loss landscape to understand the structural implications of the low-loss paths we have discovered. Second, we inspect how the construction and regularization of these curves directly influence downstream generalization performance.
%
To this end, we evaluate our Anchored (ALC) and Free (FLC) LoRA-Curve frameworks against single-mode baselines (MAP), the VI approach BLoB \citep{wangBLoB2024}, and discrete multi-mode approaches like Deep Ensembles (DE). 
We fine-tune Qwen 2.5\,7B \cite{yang2024qwen2} via LoRA across standard reasoning and classification benchmarks (ARC-Challenge, ARC-Easy \citep{clark2018arc}, OBQA \citep{mihaylov2018obqa}, boolq \citep{clark2019boolq}, Winogrande (Small/Medium)\citep{sakaguchi2021winogrande}). 
We select this scale to align with prior work \citep{samplawskiScalable2025} and avoid capacity saturation. Larger models already handle these benchmarks confidently, leaving little epistemic uncertainty for LoRA fine-tuning to capture.

Details on hyperparameter choices and our training routine can be found in \cref{sec:app_hparam-selection}, a performance comparison in \cref{sec:app_performance}, and further experimental results in \cref{sec:app_add_results}. 


\subsection{Properties of the LoRA loss-landscape}

\begin{figure}[ht]
    \centering
    \includegraphics[width=1.\linewidth]{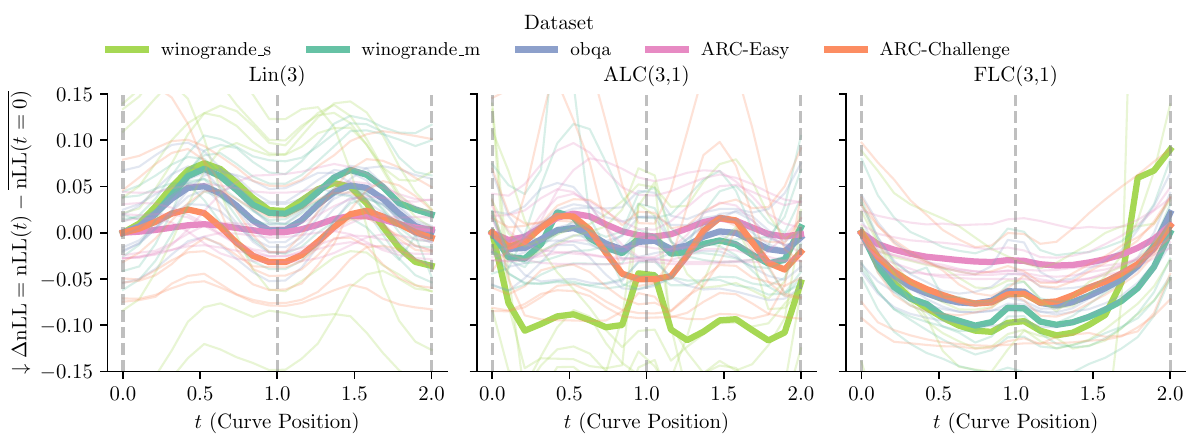}
    \caption{Mode connectivity comparison of linear vs.\ segmented curves visualized by changes in the negative log-likelihood ($\Delta$ nLL) on training data relative to the starting point along the interpolation path. 
    Vertical dashed lines denote the locations of the anchors. 
    Thick solid lines represent the mean across eight repetitions for each dataset (colors), and faded lines show individual runs.
    Although linear interpolation encounters distinct loss barriers, the ALC and especially the FLC parameterizations effectively circumvent these barriers to navigate continuous, low-loss valleys.}
    \label{fig:linear_mode_connect}
\end{figure}

To better understand the geometry of the LoRA space, we first examine whether independent optima can be connected via low-loss paths. \cref{fig:linear_mode_connect} compares linear interpolation against our segmented curve approaches. Linear interpolation (Linear(3)) encounters distinct loss barriers, manifesting as twin peaks midway between each segment at $t=0.5$ and $t=1.5$. Although the anchored curve (ALC(3,1)) reduces these barriers, strictly connecting fixed, pretrained anchors remains challenging because they may lie in suboptimal positions for continuous traversal. By contrast, the free LoRA-Curve (FLC(3,1)) easily navigates these valleys on the training data. By jointly optimizing all control points, FLC dynamically shifts the anchors to more favorable locations, thereby constructing smoother, low-loss connections. 
The overall improved flatness along the interior of the FLC path demonstrates that the LoRA loss landscape between different optima is inherently non-convex, necessitating flexible parameterizations beyond simple linear interpolation.

\begin{figure}[ht]
    \centering
    \includegraphics[width=1.\linewidth]{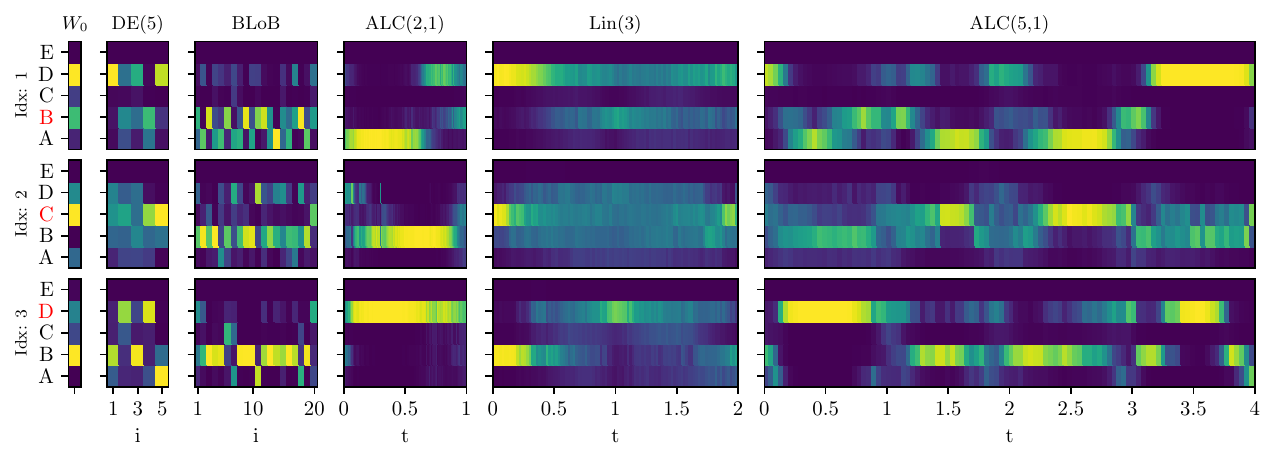}
    \caption{Class probability evolution across discrete and continuous methods, using a linear color scale where dark blue represents a probability of $0$ and yellow represents $1$ for the five labels A--E. Different rows correspond to three test examples from the ARC-Challenge dataset. Red y-axis labels indicate the ground-truth class. Facet columns represent different approaches with the base Qwen 2.5 model without LoRA pretraining, denoted as $W_0$. 
    The x-axis represents either the discrete model index $i$ or the continuous curve parameter $t$. 
    }
    \label{fig:logits_smootheness}
\end{figure}

\Cref{fig:logits_smootheness} illustrates the continuity property established in \cref{prop:continuity_main}. Unlike DE(5) and BLoB, which exhibit discrete jumps at every mode or between samples from their variational distributions, our curve parameterization guarantees smooth transitions in logit space. This provides a principled mechanism for linking continuous parameter-space traversal to functional diversity, while intermediate models still remain highly performant. This continuity is particularly striking when comparing DE(5) to ALC(5,1). Because ALC(5,1) connects these exact modes (e.g., the curve at $t=0$ corresponds to the first ensemble member $i=1$), it bridges the predictive gaps: while isolated DE members offer no insight into the functional space between them, ALC explicitly reveals the smooth transitions bridging these optima. 

Beyond just smoothness, our segmented B\'ezier parameterization leverages intermediate control points to actively inject functional diversity, in stark contrast to the rigid interpolation of linear mode connectivity. This is visually evident again in \cref{fig:logits_smootheness}, where the ALC(5,1) class probabilities shift dynamically at each new intermediate control point (i.e., at every $0.5$ increment of $t$). Consequently, curves parameterized with a higher density of control points inherently generate richer functional diversity. We quantitatively validate this in \cref{fig:plot4}.

\subsection{Implications of low-loss valleys design on the performance}

\begin{figure}[htbp]
    \centering
    \includegraphics[width=1.\linewidth]{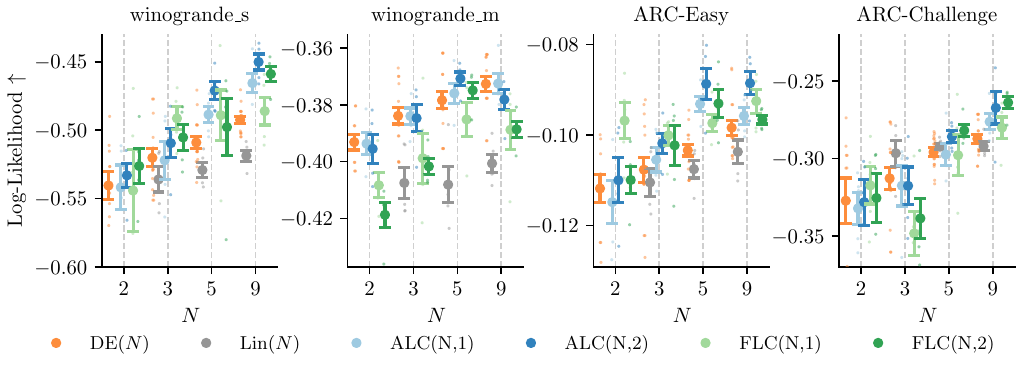}
    \caption{Comparison of our LoRA-Curve framework against DEs using test log-likelihood ($\uparrow$) across four datasets (panels) evaluated for varying numbers of anchors ($N$, x-axis). Colors distinguish discrete DE (orange) from our continuous curve parameterizations. Solid points with error bars denote the mean and standard error across 5 random seeds, shown as faded dots. 
    }
    \label{fig:performance_DE_vs_Curve}
\end{figure}

\Cref{fig:performance_DE_vs_Curve} presents a direct performance comparison between our curve approaches and discrete DEs. Because ALC uses the exact same anchors as DE, it provides a controlled 
framework that isolates both the effect of continuous versus discrete averaging (comparing  $\text{DE}(N)$ to  $\text{Lin}(N)$) and the necessity of non-linear routing (comparing $\text{ALC}(N, m)$ to $\text{Lin}(N)$).
The linear formulation (ALC with $m=0$) proves too restrictive, failing to capture the curvature of the underlying low-loss manifold. By contrast, the higher-order $\text{ALC}(N, 2)$ successfully matches or outperforms DE across datasets. 

FLC, however, exhibits less consistent gains, struggling particularly on the \texttt{winogrande\_m} dataset. This performance gap is intriguing because FLC and ALC possess identical representational capacity; in principle, FLC can represent any curve that ALC can. The discrepancy, therefore, stems strictly from the optimization strategy. While jointly optimizing all control points makes FLC significantly faster to train overall (cf.~\cref{tab:runtimes}), it induces fundamentally different optimization dynamics compared to the sequential ALC process. 
%
Furthermore, this exposes a prominent generalization gap. Although \cref{fig:linear_mode_connect} demonstrates that FLC discovers smooth, low-loss valleys in the training data, this does not consistently translate into superior test performance, indicating a misalignment between the training and test landscapes (cf.~\cref{fig:app_linear_mode_connect_traintest}).
Despite these optimization challenges, increasing the number of anchors $N$ generally improves performance for all methods (DE, ALC, and FLC). This suggests that multiple modes, as captured by longer paths, result in better performance regardless of the specific training regimen.

\begin{figure}[htbp]
    \centering
    
    
    \begin{subfigure}[b]{0.25\textwidth}
        \centering
        \includegraphics[width=1.\textwidth]{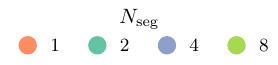} 
    \end{subfigure}
    \hfill
    \begin{subfigure}[b]{0.23\textwidth}
        \centering
        \includegraphics[width=1.\textwidth]{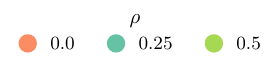}
    \end{subfigure}
    \hfill
    \begin{subfigure}[b]{0.45\textwidth}
        \centering
        \includegraphics[width=.9\textwidth]{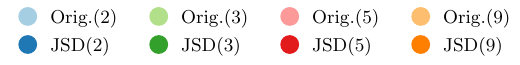}
    \end{subfigure}
    
    \vspace{-0.05cm} 
    
    
    \begin{subfigure}[b]{0.25\textwidth}
        \includegraphics[width=\textwidth]{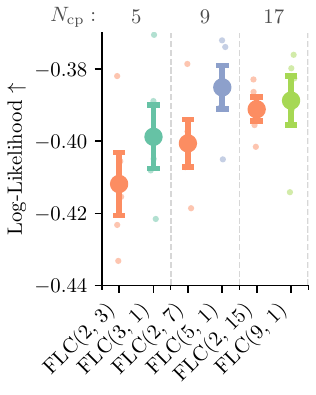}
        \vspace{-0.5cm}
        \caption{Curve vs.\ Segments}
        \label{fig:plot1}
    \end{subfigure}
    \hfill 
    \begin{subfigure}[b]{0.25\textwidth}
        \includegraphics[width=\textwidth]{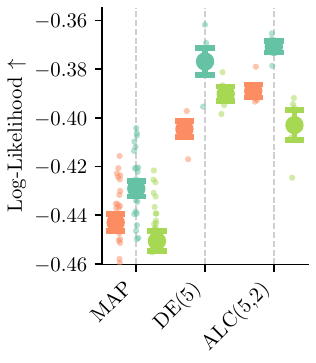}
        \caption{Flat Curve}
        \label{fig:plot2}
    \end{subfigure}
    \hfill 
    \begin{subfigure}[b]{0.24\textwidth}
        \includegraphics[width=\textwidth]{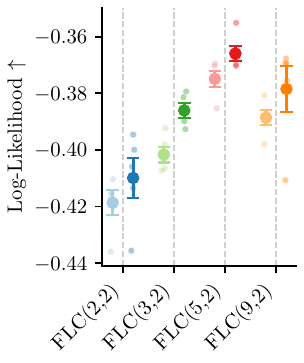}
        \caption{JSD (LL)}
        \label{fig:plot3}
    \end{subfigure} 
    \begin{subfigure}[b]{0.24\textwidth}
        \includegraphics[width=\textwidth]{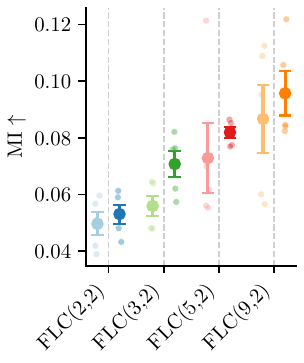}
        \caption{JSD (MI)}
        \label{fig:plot4}
    \end{subfigure}
    
    \caption{Performance and diversity analysis on the \texttt{winogrande\_m} dataset. \textbf{(a) Curve vs.\ segments:} Test log-likelihood ($\uparrow$) comparison of single-segment curves (orange) versus multi-segment curves (blue and greens), grouped by an equivalent number of control points $N_{\text{cp}}$. \textbf{(b) Flat Curve:} The effect of Flat LoRA weight perturbations ($\rho>0$, greens) compared to the unperturbed baseline ($\rho=0$, orange) for different approaches. \textbf{(c) \& (d) JSD Regularization:} The impact of JSD regularization on log-likelihood and MI ($\uparrow$), respectively. Darker shades denote JSD-regularized models, and lighter shades represent the unregularized (Orig.) baselines across varying FLC complexities. Across all panels, solid points with error bars indicate the mean and standard error, and faded dots represent individual runs.}
    \label{fig:performance_curve_properties}
\end{figure}

\Cref{fig:plot1} examines the necessity of segmentation by comparing single-segment curves against multi-segment curves, grouped by an equivalent total number of control points $N_{\text{cp}}$. Multi-segment configurations consistently match or outperform their single-segment counterparts. Furthermore, this performance advantage becomes even more pronounced as the total number of control points increases (cf.\ \cref{fig:plot1,fig:app_seg_vs_single_curve}).

\Cref{fig:plot2} investigates whether the discovered low-loss paths also benefit from wider valleys. By combining our curve framework with Flat LoRA weight perturbations ($\rho>0$), we demonstrate that optimization objectives targeting flatter regions successfully force continuous paths to lie in broader regions of the loss landscape. Consequently, the generalization benefits of locating these flatter optima---as observed in the MAP and DE baselines---transfer directly to our continuous curve framework. However, regardless of the underlying method, excessive noise injection (e.g., $\rho=0.5$) ultimately masks the training signal, indicating that the optimal perturbation strength requires careful tuning.

Finally, \cref{fig:plot3,fig:plot4} evaluate the impact of JSD regularization on performance and functional diversity. As detailed in \cref{subsec:jsd_math}, this regularizer explicitly forces the continuous path to explore functionally distinct regions. \Cref{fig:plot3} shows that JSD-regularized models (darker shades) maintain, or even improve, log-likelihoods compared to their unregularized baselines (lighter shades). Furthermore, \cref{fig:plot4} demonstrates that mutual information (MI; cf.~\cref{appendix:mi_computation}) increases through two compounding mechanisms: the addition of more anchor points $N$ and the explicit application of JSD regularization. 

\section{Conclusion}
\label{sec:conclusion}

In this work, we introduced \emph{LoRA-Curve}, a segmented B\'ezier parameterization designed to investigate the continuous posterior geometry of LLMs fine-tuned via low-rank adaptation. Contrary to recent suggestions that the LoRA landscape lacks beneficial multi-mode structure, our findings demonstrate that the space between independent optima is a rich, navigable region of low loss. 
Our free (FLC) and anchored (ALC) configurations successfully construct meaningful, continuous low-loss valleys that bypass steep loss barriers.
Furthermore, these two variants offer valuable practical trade-offs: FLC provides an efficient optimization approach by dynamically discovering valleys from scratch, whereas ALC enables the explicit, targeted connection of independently fine-tuned optima.

Navigating these non-convex valleys captures unique functional diversity. Unlike local variational methods that explore only a single basin, or discrete DEs that aggregate isolated point estimates, \emph{LoRA-Curve} guarantees smooth predictive transitions and explicitly links continuous parameter-space traversal to measurably higher mutual information without sacrificing predictive performance.
For probabilistic LoRA applications, our results establish that the continuous space between modes serves as a viable manifold for BMA, with functional diversity scaling naturally alongside the number of curve segments.
Combining this parameterization with flat-minima perturbations broadens the loss valley occupied by the curve, while JSD regularization further increases functional diversity along it.

\paragraph{Limitations}
While the \emph{LoRA-Curve} framework provides a principled approach to continuous mode connectivity, it presents certain structural and practical limitations. First, we restrict our analysis to one-dimensional, globally coupled trajectories. Extending this parameterization to higher-dimensional surfaces or assigning independent curves to individual LoRA adapters would severely degrade interpretability; our preliminary experiments also confirmed that the latter approach yields no empirical performance gains. Furthermore, while traditional BMA over continuous manifolds faces a significant computational bottleneck, we established that computing posterior weights on the training dataset in our setting leads to detrimental double-conditioning. 
Consequently, our reliance on uniform weighting ($T \to \infty$) is a practical choice that is empirically validated and motivated by a double-counting argument (see \cref{sec:appendix_temperature}), conveniently bypassing this bottleneck in our one-dimensional case.
However, in higher-dimensional parameterizations, uniform weighting would likely prove insufficient, thereby reintroducing the prohibitive cost of evaluating the full posterior. Finally, while the Free configuration (FLC) accelerates optimization, it currently exhibits a prominent generalization gap.


\bibliographystyle{plainnat}
\bibliography{references}

@inproceedings{blundellWeight2015,
  title = {Weight {{Uncertainty}} in {{Neural Network}}},
  booktitle = {Proceedings of the 32nd {{International Conference}} on {{Machine Learning}}},
  author = {Blundell, Charles and Cornebise, Julien and Kavukcuoglu, Koray and Wierstra, Daan},
  year = 2015,
  month = jun,
  pages = {1613--1622},
  publisher = {PMLR},
  issn = {1938-7228},
  urldate = {2026-04-14},
  langid = {english}
}

@inproceedings{daxbergerLaplace2021,
  title = {Laplace {{Redux}} - {{Effortless Bayesian Deep Learning}}},
  booktitle = {Advances in {{Neural Information Processing Systems}}},
  author = {Daxberger, Erik and Kristiadi, Agustinus and Immer, Alexander and Eschenhagen, Runa and Bauer, Matthias and Hennig, Philipp},
  year = 2021,
  volume = {34},
  pages = {20089--20103},
  publisher = {Curran Associates, Inc.},
  urldate = {2026-04-14}
}

@inproceedings{doldPaths2025,
  title = {Paths and {{Ambient Spaces}} in {{Neural Loss Landscapes}}},
  booktitle = {Proceedings of {{The}} 28th {{International Conference}} on {{Artificial Intelligence}} and {{Statistics}}},
  author = {Dold, Daniel and Kobialka, Julius and Palm, Nicolai and Sommer, Emanuel and R{\"u}gamer, David and D{\"u}rr, Oliver},
  year = 2025,
  month = apr,
  pages = {10--18},
  publisher = {PMLR},
  issn = {2640-3498},
  urldate = {2026-03-18},
  langid = {english}
}

@inproceedings{garipovLoss2018,
  title = {Loss {{Surfaces}}, {{Mode Connectivity}}, and {{Fast Ensembling}} of {{DNNs}}},
  booktitle = {Advances in {{Neural Information Processing Systems}}},
  author = {Garipov, Timur and Izmailov, Pavel and Podoprikhin, Dmitrii and Vetrov, Dmitry P and Wilson, Andrew G},
  year = 2018,
  volume = {31},
  publisher = {Curran Associates, Inc.},
  urldate = {2022-02-03}
}

@inproceedings{huLoRA2022,
  title = {{{LoRA}}: {{Low-Rank Adaptation}} of {{Large Language Models}}},
  booktitle = {{{ICLR}} 2022},
  author = {Hu, Edward J. and Shen, Yelong and Wallis, Phillip and {Allen-Zhu}, Zeyuan and Li, Yuanzhi and Wang, Shean and Wang, Lu and Chen, Weizhu},
  year = 2022,
  month = apr
}

@inproceedings{izmailovSubspace2020,
  title = {Subspace {{Inference}} for {{Bayesian Deep Learning}}},
  booktitle = {Proceedings of {{The}} 35th {{Uncertainty}} in {{Artificial Intelligence Conference}}},
  author = {Izmailov, Pavel and Maddox, Wesley J. and Kirichenko, Polina and Garipov, Timur and Vetrov, Dmitry and Wilson, Andrew Gordon},
  year = 2020,
  month = aug,
  pages = {1169--1179},
  publisher = {PMLR},
  issn = {2640-3498},
  urldate = {2022-10-17},
  langid = {english}
}

@inproceedings{lakshminarayananSimple2017,
  title = {Simple and {{Scalable Predictive Uncertainty Estimation}} Using {{Deep Ensembles}}},
  booktitle = {Advances in {{Neural Information Processing Systems}}},
  author = {Lakshminarayanan, Balaji and Pritzel, Alexander and Blundell, Charles},
  year = 2017,
  volume = {30},
  publisher = {Curran Associates, Inc.},
  urldate = {2026-04-14}
}

@inproceedings{liFlatLoRA2025,
  title = {Flat-{{LoRA}}: {{Low-Rank Adaptation}} over a {{Flat Loss Landscape}}},
  shorttitle = {Flat-{{LoRA}}},
  booktitle = {Proceedings of the 42nd {{International Conference}} on {{Machine Learning}}},
  author = {Li, Tao and He, Zhengbao and Li, Yujun and Wang, Yasheng and Shang, Lifeng and Huang, Xiaolin},
  year = 2025,
  month = oct,
  pages = {34549--34563},
  publisher = {PMLR},
  issn = {2640-3498},
  urldate = {2026-04-13},
  langid = {english}
}

@inproceedings{ranganathBlack2014,
  title = {Black {{Box Variational Inference}}},
  booktitle = {Proceedings of the {{Seventeenth International Conference}} on {{Artificial Intelligence}} and {{Statistics}}},
  author = {Ranganath, Rajesh and Gerrish, Sean and Blei, David},
  year = 2014,
  month = apr,
  pages = {814--822},
  publisher = {PMLR},
  issn = {1938-7228},
  urldate = {2026-04-14},
  langid = {english}
}

@inproceedings{samplawskiScalable2025,
  title = {Scalable {{Bayesian Low-Rank Adaptation}} of {{Large Language Models}} via {{Stochastic Variational Subspace Inference}}},
  booktitle = {Conference on {{Uncertainty}} in {{Artificial Intelligence}}},
  author = {Samplawski, Colin and Cobb, Adam D and Acharya, Manoj and Kaur, Ramneet and Jha, Susmit},
  year = 2025,
  pages = {3587--3604},
  publisher = {PMLR}
}

@article{wangBLoB2024,
  title = {{{BLoB}}: {{Bayesian Low-Rank Adaptation}} by {{Backpropagation}} for {{Large Language Models}}},
  shorttitle = {{{BLoB}}},
  author = {Wang, Yibin and Shi, Haizhou and Han, Ligong and Metaxas, Dimitris and Wang, Hao},
  year = 2024,
  month = dec,
  journal = {Advances in Neural Information Processing Systems},
  volume = {37},
  pages = {67758--67794},
  doi = {10.52202/079017-2164},
  urldate = {2025-12-28},
  langid = {english}
}

@inproceedings{balabanovUncertainty2025,
  title = {Uncertainty Quantification in Fine-Tuned {{LLMs}} Using {{LoRA}} Ensembles},
  booktitle = {{{ICLR Workshop}}: {{Quantify Uncertainty}} and {{Hallucination}} in {{Foundation Models}}: {{The Next Frontier}} in {{Reliable AI}}},
  author = {Balabanov, Oleksandr and Linander, Hampus},
  year = 2025
}

@inproceedings{yangBayesian2024,
  title = {Bayesian {{Low-rank Adaptation}} for {{Large Language Models}}},
  booktitle = {The {{Twelfth International Conference}} on {{Learning Representations}}},
  author = {Yang, Adam X. and Robeyns, Maxime and Wang, Xi and Aitchison, Laurence},
  year = 2024
}

@inproceedings{
li2025calibrating,
title={Calibrating {LLM}s with Information-Theoretic Evidential Deep Learning},
author={Yawei Li and David R{\"u}gamer and Bernd Bischl and Mina Rezaei},
booktitle={The Thirteenth International Conference on Learning Representations},
year={2025},
OPTurl={https://openreview.net/forum?id=YcML3rJl0N}
}

@article{houlsby2011bayesian,
  title={Bayesian active learning for classification and preference learning},
  author={Houlsby, Neil and Husz{\'a}r, Ferenc and Ghahramani, Zoubin and Lengyel, M{\'a}t{\'e}},
  journal={arXiv preprint arXiv:1112.5745},
  year={2011}
}

@inproceedings{draxler2018essentially,
  title={Essentially no barriers in neural network energy landscape},
  author={Draxler, Felix and Veschgini, Kambis and Salmhofer, Manfred and Hamprecht, Fred},
  booktitle={International conference on machine learning},
  pages={1309--1318},
  year={2018},
  organization={PMLR}
}

@article{yang2024qwen2,
  title={Qwen2. 5 Technical Report},
  author={Yang, An and Yang, Baosong and Zhang, Beichen and Hui, Binyuan and Zheng, Bo and Yu, Bowen and Li, Chengyuan and Liu, Dayiheng and Huang, Fei and Wei, Haoran and others},
  journal={arXiv preprint arXiv:2412.15115},
  year={2024}
}

@article{sakaguchi2021winogrande,
  title={Winogrande: An adversarial winograd schema challenge at scale},
  author={Sakaguchi, Keisuke and Bras, Ronan Le and Bhagavatula, Chandra and Choi, Yejin},
  journal={Communications of the ACM},
  volume={64},
  number={9},
  pages={99--106},
  year={2021},
  publisher={ACM New York, NY, USA}
}

@article{clark2018arc,
  title={Think you have solved question answering? try arc, the ai2 reasoning challenge},
  author={Clark, Peter and Cowhey, Isaac and Etzioni, Oren and Khot, Tushar and Sabharwal, Ashish and Schoenick, Carissa and Tafjord, Oyvind},
  journal={arXiv preprint arXiv:1803.05457},
  year={2018}
}

@inproceedings{mihaylov2018obqa,
  title={Can a suit of armor conduct electricity? a new dataset for open book question answering},
  author={Mihaylov, Todor and Clark, Peter and Khot, Tushar and Sabharwal, Ashish},
  booktitle={Proceedings of the 2018 conference on empirical methods in natural language processing},
  pages={2381--2391},
  year={2018}
}

@inproceedings{clark2019boolq,
  title={Boolq: Exploring the surprising difficulty of natural yes/no questions},
  author={Clark, Christopher and Lee, Kenton and Chang, Ming-Wei and Kwiatkowski, Tom and Collins, Michael and Toutanova, Kristina},
  booktitle={Proceedings of the 2019 conference of the north American chapter of the association for computational linguistics: Human language technologies, volume 1 (long and short papers)},
  pages={2924--2936},
  year={2019}
}


\appendix

\clearpage
\section{Mutual information (MI)}
\label{appendix:mi_computation}
To quantify epistemic uncertainty as a measure of functional diversity within our LoRA-Curve framework, we compute the Mutual Information (MI) between the curve parameter $t$ and the predictive output $y$, following the Bayesian Active Learning by Disagreement (BALD) objective \citep{houlsby2011bayesian}. Assuming a distribution $p(t)$ over the curve parameter $t \in [0, T_{\max}]$, the theoretical continuous MI for a single input $\mathbf{x}$ is defined as:
$$ I[y, t \mid \mathbf{x}, \mathcal{D}] \equiv \text{MI}(\mathbf{x}) = \text{H} \left[ \mathbb{E}_{p(t)} [p(y \mid \mathbf{x}, \theta_\vartheta(t))] \right] - \mathbb{E}_{p(t)} \left[ \text{H} [p(y \mid \mathbf{x}, \theta_\vartheta(t))] \right] $$
where $\theta_\vartheta(t)$ denotes the model parameters at curve position $t$.

In practice, we approximate this continuous objective via numerical integration:
$$ \text{MI}(\mathbf{x}) \approx \underbrace{\text{H} \left[ \sum_{j=1}^M w_j \, p(y \mid \mathbf{x}, \theta_\vartheta(t_j)) \right]}_{\text{Total Uncertainty}} - \underbrace{\sum_{j=1}^M w_j \, \text{H} [p(y \mid \mathbf{x}, \theta_\vartheta(t_j))]}_{\text{Expected Data Uncertainty}} $$
where $\{t_j\}_{j=1}^M$ represents $M$ equispaced grid points along the curve (typically $M = 20$) and $w_j$ represent numerical quadrature weights (defaulting to uniform weights $w_j = 1/M$ for a uniform posterior $p(t|\mathcal{D})$).

In our implementation, we evaluate this objective by first calculating the total predictive uncertainty across the entire curve. This is achieved by computing the weighted average predictive distribution $\bar{p} = \sum_{j=1}^M w_j \, p_j$, where $p_j = \operatorname{softmax}(f(\mathbf{x}; \theta_\vartheta(t_j)))$, and evaluating its entropy $H[\bar{p}]$. We then capture the expected pointwise (aleatoric) uncertainty by computing the weighted average of the individual prediction entropies, $\sum_j w_j H[p_j]$. The epistemic uncertainty (the MI score) is subsequently defined as the difference between these two quantities: $H[\bar{p}] - \sum_j w_j H[p_j]$. Consequently, a high MI score indicates that predictions vary significantly along the curve---reflecting high functional disagreement between different $t$ values---even if individual parameter configurations remain confident. This effectively isolates the functional diversity captured by the continuous trajectory.

Finally, the single scalar MI metric reported in our experiments represents the mean mutual information computed across all datapoints in the evaluation dataset $\mathcal{D}$:
$$ \text{MI}_{\text{reported}} = \frac{1}{|\mathcal{D}|} \sum_{\mathbf{x} \in \mathcal{D}} \text{MI}(\mathbf{x}) $$

\section{Temperature scaling and Bayesian model averaging}
\label{sec:appendix_temperature}

In our setting, Bayesian model averaging (BMA) requires evaluating the posterior probability at discrete grid points $\{t_i\}_{i=1}^N$ on the training dataset $\mathcal{D}$. We formulate this using a grid approximation over a uniform prior $p(t) \sim U(0, \text{seg})$ with a temperature scaling parameter $T$:
$$
    p(y^\ast \mid x^\ast, \mathcal{D}) \approx \sum_{j=1}^{M} \varpi_j(T)\, p\left(y^\ast \mid x^\ast, W_{\vartheta}(t_j), \mathcal{D}\right), 
    \quad \varpi_j(T) = \frac{p(\mathcal{D} \mid t_j)^{1/T}}{\sum_{k=1}^{M} p(\mathcal{D} \mid t_k)^{1/T}},
$$

Optimizing the temperature on a validation set reveals that uniform averaging---corresponding to an infinite temperature ($T \to \infty$)---yields the best empirical performance. This is justified mathematically by the limit:
$$\lim_{T \to \infty} \varpi_i(T) = \lim_{T \to \infty} \frac{p(\mathcal{D} \mid t_i)^{1/T} p(t_i)}{\sum_{k=1}^N p(\mathcal{D} \mid t_k)^{1/T} p(t_k)} = \frac{1}{N}.$$

\begin{figure}[htb]
    \centering
    \includegraphics[width=1.\linewidth]{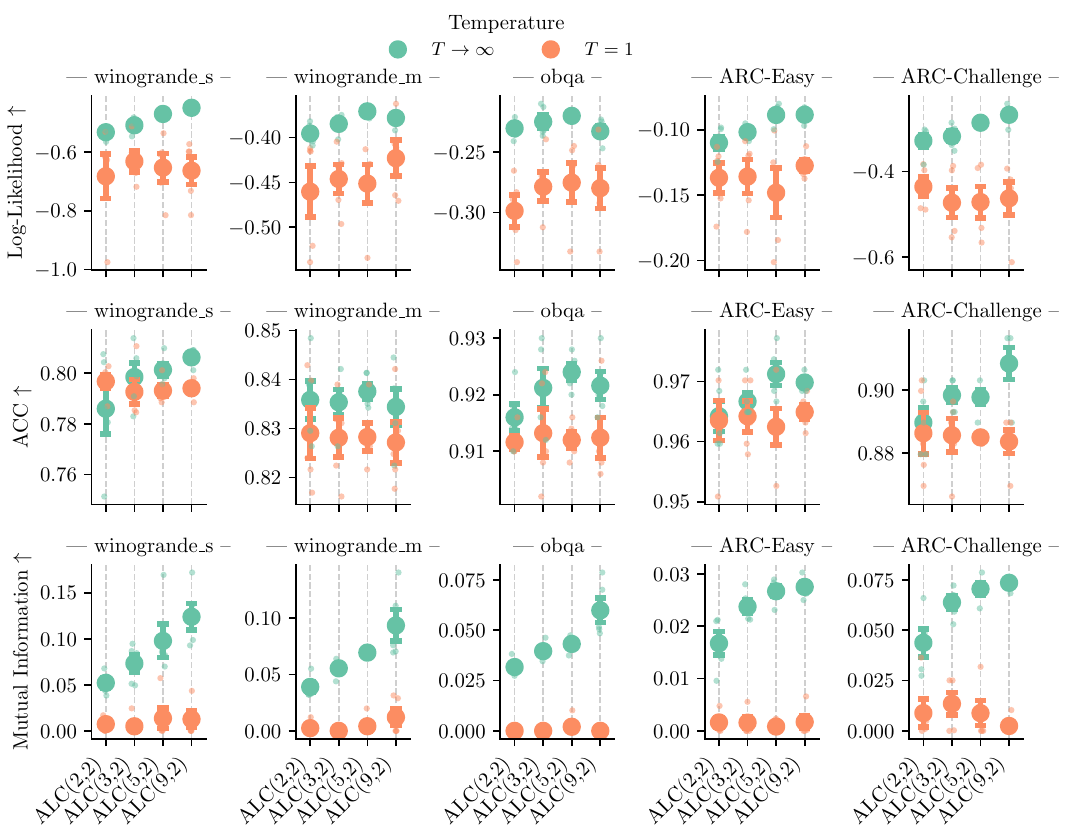}
    \caption{Effect of temperature scaling on Bayesian model averaging. Performance comparison of standard posterior weighting ($T=1$, orange) versus uniform averaging ($T \to \infty$, green) across five datasets (columns). The rows display test Log-Likelihood ($\uparrow$), Accuracy (ACC $\uparrow$), and Mutual Information ($\uparrow$) for varying Anchored LoRA-Curve (ALC) complexities. Solid points with error bars indicate the mean and standard error, while faded dots represent individual runs.}
    \label{fig:app_ppd_vs_uniform}
\end{figure}

Because the curve optimization already maximizes data likelihood along the curve, computing a standard posterior without temperature scaling ($T=1$) heavily over-prioritizes the data contribution. This double-counting leads to overconfident predictions and degrades the model's performance back to single-point estimates (cf.~\cref{fig:app_ppd_vs_uniform}). As demonstrated across all metrics, this uniform averaging effectively mitigates overconfidence, consistently outperforming $T=1$ and yielding superior empirical performance and functional diversity.
Consequently, we discard standard temperature scaling and approximate the Bayesian predictive distribution via the uniform averaging ($T \to \infty$) defined in the main text.

\section{Performance comparison}
\label{sec:app_performance}

\Cref{tab:performance_metrics_updated} compares MAP, BLoB \cite{wangBLoB2024}, ScalaBL \cite{samplawskiScalable2025}, and DE \cite{lakshminarayananSimple2017} against our curve parameterizations. We reproduced the BLoB and ScalaBL baselines using the exact framework released by \citet{samplawskiScalable2025} across seeds $(0, 1, \dots, 7)$. The DE and MAP results were extracted directly within our primary ALC experiments, as $\mathrm{ALC}(N,m)$ intrinsically encapsulates both $\mathrm{DE}(N)$ anchors and $N$ individual MAP solutions.

\begin{table*}[htbp]
\centering
\caption{Performance comparison of Bayesian Model Averaging (BMA) strategies. Evaluation of Accuracy (ACC $\uparrow$), Expected Calibration Error (ECE $\downarrow$), and Log-Likelihood (LL $\uparrow$) across five datasets. $N$ denotes the number of anchor points, and $m$ represents the curve segment degree ($m=0$: discrete/linear, $m>0$: continuous curve). ALC (Anchored LoRA-Curve) uses frozen anchors from pretrained MAP solutions, while FLC (Free LoRA-Curve) jointly optimizes all control points. Results report the mean $\pm$ $2\sigma$ across available seeds. The best metric per dataset is \textbf{bolded}, and models whose mean $\pm$ $2\sigma$ interval contains the best result are \underline{underlined}. Notably, our continuous curve approaches (ALC and FLC) perform competitively with, and frequently outperform, discrete deep ensembles (DE) and baseline methods.}
\label{tab:performance_metrics_updated}
\resizebox{\textwidth}{!}{%
\begin{tabular}{llcccccccc}
\toprule
\textbf{Metric} & \textbf{Model} & \textbf{N} & \textbf{m} & \textbf{ARC-Challenge} & \textbf{ARC-Easy} & \textbf{obqa} & \textbf{winogrande\_m} & \textbf{winogrande\_s} & \textbf{boolq} \\
\midrule
\multirow{13}{*}{ACC $\uparrow$} & MAP & 1 & 0 & \underline{0.889 $\pm$ 0.021} & 0.962 $\pm$ 0.009 & \underline{0.910 $\pm$ 0.016} & \underline{0.820 $\pm$ 0.018} & \underline{0.779 $\pm$ 0.030} & \underline{0.892 $\pm$ 0.009} \\
 & BLoB & --- & --- & \underline{0.907 $\pm$ 0.010} & 0.965 $\pm$ 0.006 & \underline{0.917 $\pm$ 0.013} & \underline{0.828 $\pm$ 0.011} & 0.788 $\pm$ 0.012 & 0.891 $\pm$ 0.004 \\
 & ScalaBL & --- & --- & \underline{0.902 $\pm$ 0.012} & 0.963 $\pm$ 0.003 & 0.912 $\pm$ 0.009 & 0.817 $\pm$ 0.007 & 0.787 $\pm$ 0.009 & 0.883 $\pm$ 0.005 \\
 & Lin(3) & 3 & 0 & \underline{0.893 $\pm$ 0.021} & 0.966 $\pm$ 0.005 & \underline{0.914 $\pm$ 0.013} & 0.823 $\pm$ 0.009 & \underline{0.773 $\pm$ 0.040} & \underline{0.894 $\pm$ 0.009} \\
 & DE(3) & 3 & 0 & \underline{0.897 $\pm$ 0.016} & 0.964 $\pm$ 0.005 & \underline{0.920 $\pm$ 0.012} & \underline{0.835 $\pm$ 0.018} & \underline{0.787 $\pm$ 0.029} & \underline{0.895 $\pm$ 0.006} \\
 & DE(9) & 9 & 0 & \underline{0.902 $\pm$ 0.011} & \underline{0.970 $\pm$ 0.005} & \underline{0.923 $\pm$ 0.011} & \textbf{0.838 $\pm$ 0.013} & \underline{0.797 $\pm$ 0.016} & \underline{0.897 $\pm$ 0.003} \\
 & ALC(2,1) & 2 & 1 & \underline{0.896 $\pm$ 0.018} & 0.965 $\pm$ 0.006 & \underline{0.919 $\pm$ 0.008} & \underline{0.833 $\pm$ 0.018} & 0.790 $\pm$ 0.015 & \textbf{0.900 $\pm$ 0.005} \\
 & ALC(3,2) & 3 & 2 & \underline{0.898 $\pm$ 0.010} & \underline{0.967 $\pm$ 0.006} & \underline{0.921 $\pm$ 0.015} & \underline{0.835 $\pm$ 0.012} & \underline{0.798 $\pm$ 0.024} & \underline{0.899 $\pm$ 0.005} \\
 & FLC(3,2) & 3 & 2 & \underline{0.898 $\pm$ 0.015} & \underline{0.966 $\pm$ 0.009} & \underline{0.922 $\pm$ 0.006} & \underline{0.833 $\pm$ 0.017} & \underline{0.796 $\pm$ 0.019} & - \\
 & ALC(5,2) & 5 & 2 & 0.898 $\pm$ 0.010 & \textbf{0.971 $\pm$ 0.008} & \textbf{0.924 $\pm$ 0.006} & \underline{0.838 $\pm$ 0.007} & \underline{0.801 $\pm$ 0.011} & \underline{0.898 $\pm$ 0.006} \\
 & FLC(5,2) & 5 & 2 & \underline{0.903 $\pm$ 0.023} & \underline{0.971 $\pm$ 0.005} & \underline{0.923 $\pm$ 0.008} & \underline{0.835 $\pm$ 0.012} & \underline{0.801 $\pm$ 0.017} & - \\
 & ALC(9,2) & 9 & 2 & \textbf{0.908 $\pm$ 0.022} & \underline{0.970 $\pm$ 0.003} & \underline{0.922 $\pm$ 0.011} & \underline{0.834 $\pm$ 0.017} & \textbf{0.806 $\pm$ 0.007} & \underline{0.895 $\pm$ 0.008} \\
 & FLC(9,2) & 9 & 2 & \underline{0.906 $\pm$ 0.013} & \underline{0.970 $\pm$ 0.002} & \underline{0.918 $\pm$ 0.009} & 0.831 $\pm$ 0.005 & \underline{0.802 $\pm$ 0.013} & 0.876 $\pm$ 0.001 \\
\midrule
\multirow{13}{*}{ECE $\downarrow$} & MAP & 1 & 0 & \underline{0.068 $\pm$ 0.035} & \underline{0.024 $\pm$ 0.012} & \underline{0.034 $\pm$ 0.018} & 0.063 $\pm$ 0.033 & 0.103 $\pm$ 0.044 & \underline{0.023 $\pm$ 0.017} \\
 & BLoB & --- & --- & \textbf{0.034 $\pm$ 0.014} & \underline{0.015 $\pm$ 0.008} & \textbf{0.022 $\pm$ 0.009} & 0.043 $\pm$ 0.009 & 0.072 $\pm$ 0.012 & \textbf{0.015 $\pm$ 0.007} \\
 & ScalaBL & --- & --- & 0.044 $\pm$ 0.006 & 0.018 $\pm$ 0.004 & \underline{0.024 $\pm$ 0.018} & \underline{0.037 $\pm$ 0.010} & 0.086 $\pm$ 0.011 & 0.022 $\pm$ 0.003 \\
 & Lin(3) & 3 & 0 & \underline{0.047 $\pm$ 0.021} & 0.023 $\pm$ 0.011 & \underline{0.022 $\pm$ 0.011} & 0.044 $\pm$ 0.009 & 0.080 $\pm$ 0.006 & \underline{0.034 $\pm$ 0.049} \\
 & DE(3) & 3 & 0 & \underline{0.050 $\pm$ 0.023} & \underline{0.018 $\pm$ 0.007} & \underline{0.029 $\pm$ 0.007} & \underline{0.034 $\pm$ 0.017} & 0.078 $\pm$ 0.016 & \underline{0.016 $\pm$ 0.004} \\
 & DE(9) & 9 & 0 & \underline{0.046 $\pm$ 0.015} & \underline{0.017 $\pm$ 0.006} & \underline{0.025 $\pm$ 0.011} & \textbf{0.028 $\pm$ 0.006} & 0.069 $\pm$ 0.008 & \underline{0.017 $\pm$ 0.006} \\
 & ALC(2,1) & 2 & 1 & 0.061 $\pm$ 0.014 & 0.021 $\pm$ 0.005 & \underline{0.028 $\pm$ 0.019} & 0.047 $\pm$ 0.016 & 0.088 $\pm$ 0.041 & \underline{0.025 $\pm$ 0.016} \\
 & ALC(3,2) & 3 & 2 & 0.056 $\pm$ 0.021 & \underline{0.014 $\pm$ 0.009} & \underline{0.029 $\pm$ 0.014} & \underline{0.032 $\pm$ 0.023} & 0.077 $\pm$ 0.035 & \underline{0.017 $\pm$ 0.008} \\
 & FLC(3,2) & 3 & 2 & 0.060 $\pm$ 0.019 & \underline{0.015 $\pm$ 0.007} & \underline{0.026 $\pm$ 0.009} & \underline{0.049 $\pm$ 0.037} & 0.078 $\pm$ 0.025 & - \\
 & ALC(5,2) & 5 & 2 & \underline{0.043 $\pm$ 0.019} & \underline{0.016 $\pm$ 0.008} & \underline{0.024 $\pm$ 0.015} & \underline{0.033 $\pm$ 0.016} & \underline{0.054 $\pm$ 0.025} & \underline{0.022 $\pm$ 0.022} \\
 & FLC(5,2) & 5 & 2 & 0.051 $\pm$ 0.010 & \underline{0.015 $\pm$ 0.009} & \underline{0.027 $\pm$ 0.015} & \underline{0.035 $\pm$ 0.011} & \underline{0.072 $\pm$ 0.046} & - \\
 & ALC(9,2) & 9 & 2 & \underline{0.041 $\pm$ 0.016} & \textbf{0.012 $\pm$ 0.005} & \underline{0.035 $\pm$ 0.017} & \underline{0.030 $\pm$ 0.020} & \textbf{0.039 $\pm$ 0.026} & 0.038 $\pm$ 0.010 \\
 & FLC(9,2) & 9 & 2 & \underline{0.038 $\pm$ 0.018} & \underline{0.018 $\pm$ 0.008} & \underline{0.031 $\pm$ 0.021} & \underline{0.033 $\pm$ 0.018} & \underline{0.048 $\pm$ 0.032} & 0.022 $\pm$ 0.004 \\
\midrule
\multirow{13}{*}{LL $\uparrow$} & MAP & 1 & 0 & \underline{-0.393 $\pm$ 0.180} & \underline{-0.131 $\pm$ 0.050} & -0.258 $\pm$ 0.045 & -0.432 $\pm$ 0.039 & \underline{-0.583 $\pm$ 0.148} & -0.273 $\pm$ 0.020 \\
 & BLoB & --- & --- & -0.278 $\pm$ 0.013 & \underline{-0.095 $\pm$ 0.007} & \textbf{-0.201 $\pm$ 0.015} & \underline{-0.377 $\pm$ 0.018} & -0.485 $\pm$ 0.016 & \textbf{-0.225 $\pm$ 0.010} \\
 & ScalaBL & --- & --- & -0.295 $\pm$ 0.017 & -0.109 $\pm$ 0.006 & -0.224 $\pm$ 0.008 & -0.393 $\pm$ 0.013 & -0.501 $\pm$ 0.015 & -0.237 $\pm$ 0.007 \\
 & Lin(3) & 3 & 0 & \underline{-0.297 $\pm$ 0.037} & -0.111 $\pm$ 0.014 & -0.243 $\pm$ 0.022 & -0.408 $\pm$ 0.025 & -0.536 $\pm$ 0.041 & -0.273 $\pm$ 0.035 \\
 & DE(3) & 3 & 0 & \underline{-0.313 $\pm$ 0.055} & \underline{-0.108 $\pm$ 0.021} & -0.225 $\pm$ 0.018 & \underline{-0.384 $\pm$ 0.022} & -0.520 $\pm$ 0.050 & -0.259 $\pm$ 0.011 \\
 & DE(9) & 9 & 0 & -0.287 $\pm$ 0.023 & \underline{-0.098 $\pm$ 0.013} & -0.218 $\pm$ 0.013 & \underline{-0.373 $\pm$ 0.020} & -0.493 $\pm$ 0.021 & -0.253 $\pm$ 0.006 \\
 & ALC(2,1) & 2 & 1 & -0.332 $\pm$ 0.045 & -0.115 $\pm$ 0.021 & -0.230 $\pm$ 0.014 & -0.394 $\pm$ 0.018 & -0.542 $\pm$ 0.073 & -0.257 $\pm$ 0.012 \\
 & ALC(3,2) & 3 & 2 & \underline{-0.318 $\pm$ 0.054} & -0.102 $\pm$ 0.010 & \underline{-0.225 $\pm$ 0.025} & \underline{-0.385 $\pm$ 0.021} & -0.510 $\pm$ 0.047 & -0.251 $\pm$ 0.008 \\
 & FLC(3,2) & 3 & 2 & -0.339 $\pm$ 0.058 & \underline{-0.102 $\pm$ 0.020} & -0.231 $\pm$ 0.015 & -0.402 $\pm$ 0.013 & -0.505 $\pm$ 0.042 & - \\
 & ALC(5,2) & 5 & 2 & -0.286 $\pm$ 0.017 & \underline{-0.089 $\pm$ 0.015} & -0.220 $\pm$ 0.006 & \textbf{-0.371 $\pm$ 0.010} & \underline{-0.471 $\pm$ 0.030} & -0.258 $\pm$ 0.022 \\
 & FLC(5,2) & 5 & 2 & -0.282 $\pm$ 0.017 & \underline{-0.093 $\pm$ 0.014} & -0.229 $\pm$ 0.013 & \underline{-0.375 $\pm$ 0.013} & \underline{-0.498 $\pm$ 0.094} & - \\
 & ALC(9,2) & 9 & 2 & \underline{-0.267 $\pm$ 0.046} & \textbf{-0.089 $\pm$ 0.011} & -0.233 $\pm$ 0.019 & \underline{-0.378 $\pm$ 0.016} & \textbf{-0.450 $\pm$ 0.026} & -0.273 $\pm$ 0.006 \\
 & FLC(9,2) & 9 & 2 & \textbf{-0.264 $\pm$ 0.017} & -0.097 $\pm$ 0.004 & -0.236 $\pm$ 0.021 & -0.389 $\pm$ 0.012 & \underline{-0.459 $\pm$ 0.024} & -0.298 $\pm$ 0.002 \\
\midrule
\multirow{11}{*}{MI $\uparrow$} & MAP & 1 & 0 & 0.000 $\pm$ 0.000 & 0.000 $\pm$ 0.000 & 0.000 $\pm$ 0.000 & 0.000 $\pm$ 0.000 & 0.000 $\pm$ 0.000 & 0.000 $\pm$ 0.000 \\
 & Lin(3) & 3 & 0 & 0.027 $\pm$ 0.009 & 0.008 $\pm$ 0.005 & 0.017 $\pm$ 0.003 & 0.026 $\pm$ 0.005 & 0.023 $\pm$ 0.021 & 0.027 $\pm$ 0.024 \\
 & DE(3) & 3 & 0 & 0.035 $\pm$ 0.017 & 0.013 $\pm$ 0.007 & 0.023 $\pm$ 0.009 & 0.031 $\pm$ 0.009 & 0.030 $\pm$ 0.031 & 0.013 $\pm$ 0.004 \\
 & DE(9) & 9 & 0 & 0.051 $\pm$ 0.012 & 0.018 $\pm$ 0.006 & 0.033 $\pm$ 0.003 & 0.047 $\pm$ 0.009 & 0.047 $\pm$ 0.024 & 0.016 $\pm$ 0.003 \\
 & ALC(2,1) & 2 & 1 & 0.040 $\pm$ 0.012 & 0.015 $\pm$ 0.010 & 0.026 $\pm$ 0.017 & 0.034 $\pm$ 0.022 & 0.049 $\pm$ 0.027 & 0.017 $\pm$ 0.005 \\
 & ALC(3,2) & 3 & 2 & \underline{0.064 $\pm$ 0.015} & \underline{0.024 $\pm$ 0.006} & 0.040 $\pm$ 0.009 & 0.055 $\pm$ 0.015 & 0.073 $\pm$ 0.043 & 0.020 $\pm$ 0.004 \\
 & FLC(3,2) & 3 & 2 & 0.060 $\pm$ 0.013 & 0.021 $\pm$ 0.004 & 0.038 $\pm$ 0.011 & 0.056 $\pm$ 0.016 & 0.070 $\pm$ 0.021 & - \\
 & ALC(5,2) & 5 & 2 & \underline{0.071 $\pm$ 0.013} & \underline{0.027 $\pm$ 0.005} & 0.043 $\pm$ 0.008 & 0.069 $\pm$ 0.007 & \underline{0.098 $\pm$ 0.081} & \underline{0.037 $\pm$ 0.038} \\
 & FLC(5,2) & 5 & 2 & 0.061 $\pm$ 0.012 & 0.022 $\pm$ 0.002 & \underline{0.045 $\pm$ 0.018} & \underline{0.073 $\pm$ 0.055} & 0.077 $\pm$ 0.030 & - \\
 & ALC(9,2) & 9 & 2 & \textbf{0.074 $\pm$ 0.007} & \textbf{0.028 $\pm$ 0.004} & \textbf{0.060 $\pm$ 0.027} & \textbf{0.093 $\pm$ 0.063} & \textbf{0.124 $\pm$ 0.064} & \textbf{0.073 $\pm$ 0.008} \\
 & FLC(9,2) & 9 & 2 & \underline{0.065 $\pm$ 0.013} & \underline{0.027 $\pm$ 0.006} & \underline{0.051 $\pm$ 0.023} & \underline{0.087 $\pm$ 0.053} & \underline{0.107 $\pm$ 0.049} & 0.015 $\pm$ 0.003 \\
\bottomrule
\end{tabular}%
}
\end{table*}

\clearpage

\section{Additional results}
\label{sec:app_add_results}

\begin{figure}[htb]
    \centering
    \includegraphics[width=1.\linewidth]{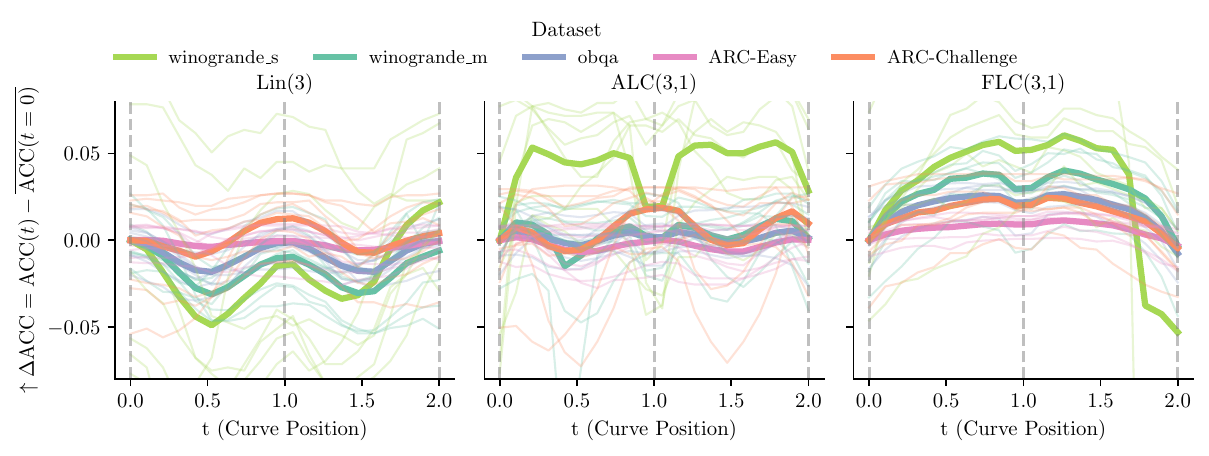}
    \caption{Same as \cref{fig:linear_mode_connect}, but evaluating the change in accuracy on y-axis}
    \label{fig:app_linear_mode_connect_acc}
\end{figure}

\begin{figure}[htb]
    \centering
    \includegraphics[width=1.\linewidth]{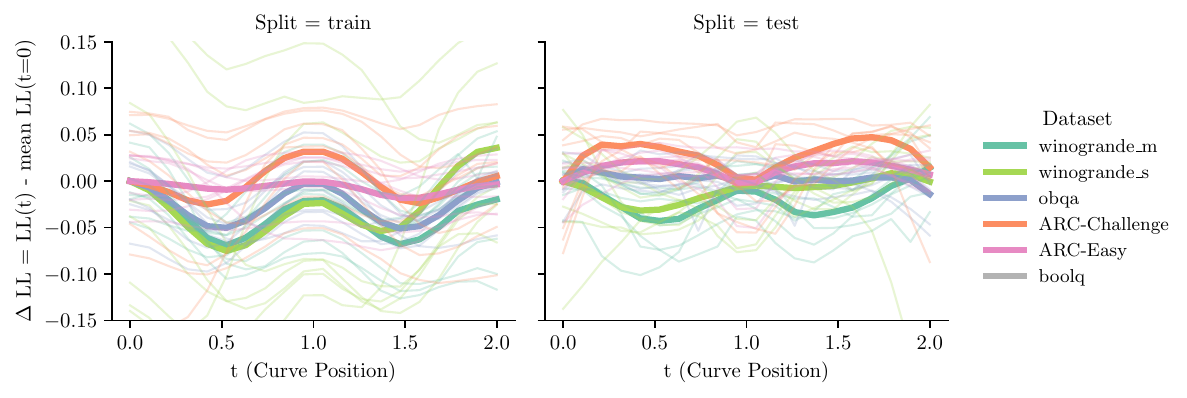}
    \caption{Linear mode connectivity for train and test splits. The left panel depicts the relative change in log-likelihood ($\Delta$ LL) for a Lin(3) model, identical to the setup in \cref{fig:linear_mode_connect}. The right panel extends this analysis to the test data, illustrating how the training loss barriers (dips at $t=0.5$ and $t=1.5$) manifest differently on the test set across various datasets.}
    \label{fig:app_linear_mode_connect_traintest}
\end{figure}

\begin{figure}[htbp]
    \centering
    \includegraphics[width=1.\linewidth]{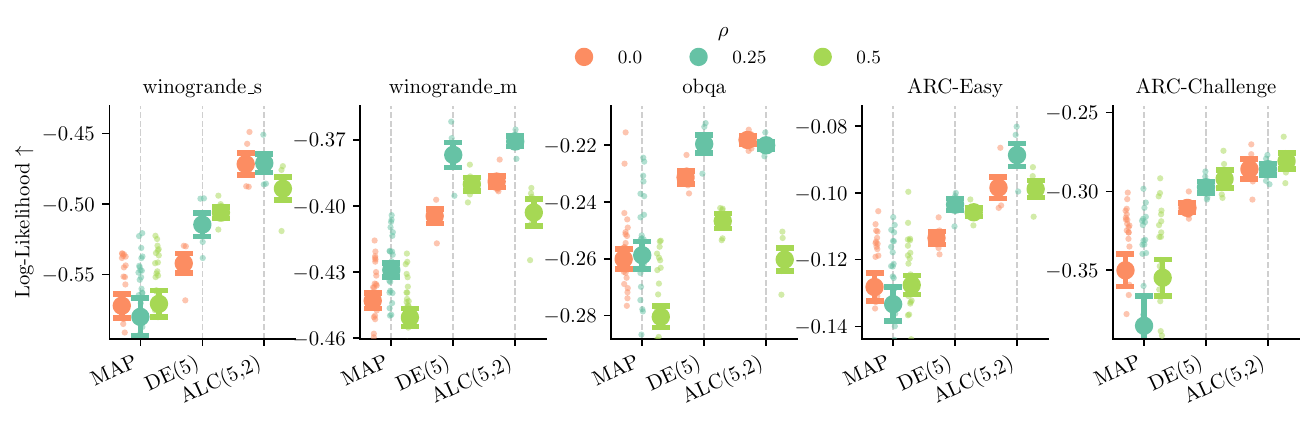}
    \caption{Same comparison as in \cref{fig:plot1}, but evaluated on the Winogrande S/M, OBQA, and ARC-Challenge/Easy datasets, which are displayed across different columns.}
    \label{fig:app_performance_flat_lora}
\end{figure}

\begin{figure}[htb]
    \centering
    \includegraphics[width=1.\linewidth]{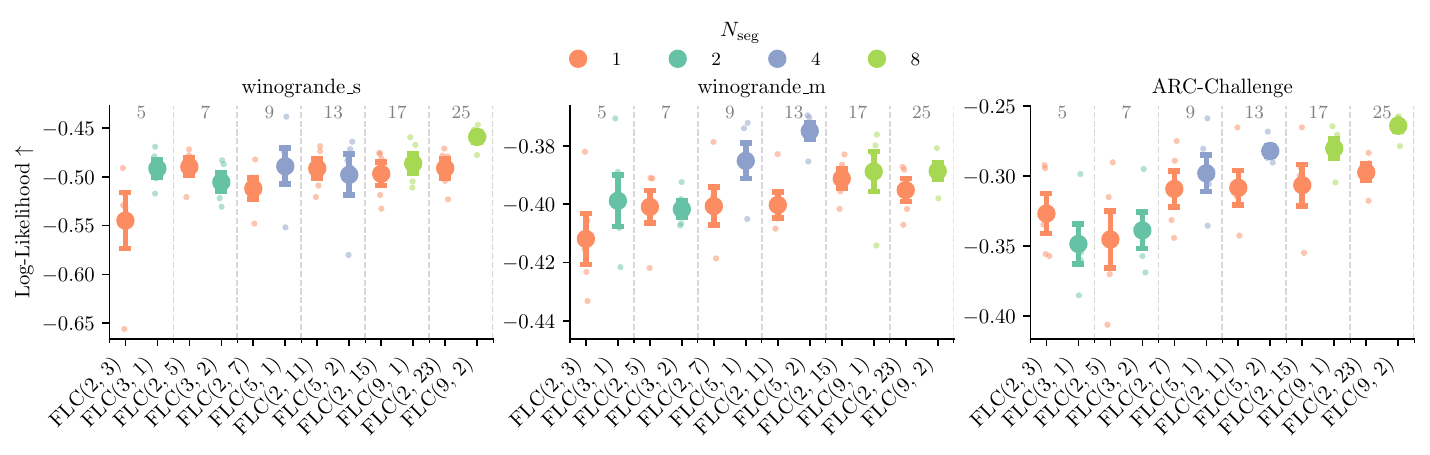}
    \caption{Log-likelihood comparison of multi-segment vs. single-segment B\'ezier curves. Test data log-likelihood ($\uparrow$) is evaluated across three datasets (panels). Models are grouped by the total number of control points, indicated by the gray numbers at the top of each section. Color coding denotes the number of segments ($N_{\text{seg}}$), with the orange baseline representing the single-segment approach. For a fixed control point budget, multi-segment curves consistently match or outperform their single-segment counterparts.}
    \label{fig:app_seg_vs_single_curve}
\end{figure}

\begin{figure}[htbp]
    \centering
    \includegraphics[width=.8\linewidth]{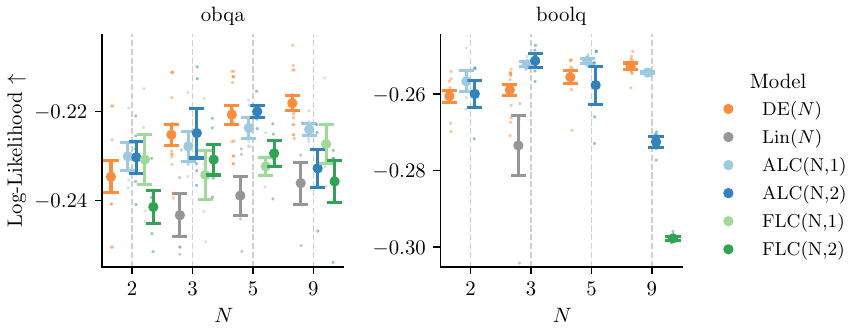}
    \caption{Extended test log-likelihood comparison. Same as \cref{fig:performance_DE_vs_Curve}, but extended to include the \texttt{obqa} and \texttt{boolq} datasets. The panels compare our continuous LoRA-Curve parameterizations against discrete Deep Ensembles (DE, orange) using test log-likelihood ($\uparrow$) evaluated for varying numbers of anchors ($N$, x-axis). Solid points with error bars denote the mean and standard error across 5 random seeds, shown as faded dots.}
    \label{fig:app_performance_DE_vs_Curve}
\end{figure}

\begin{figure}[htbp]
    \centering
    \includegraphics[width=1.\linewidth]{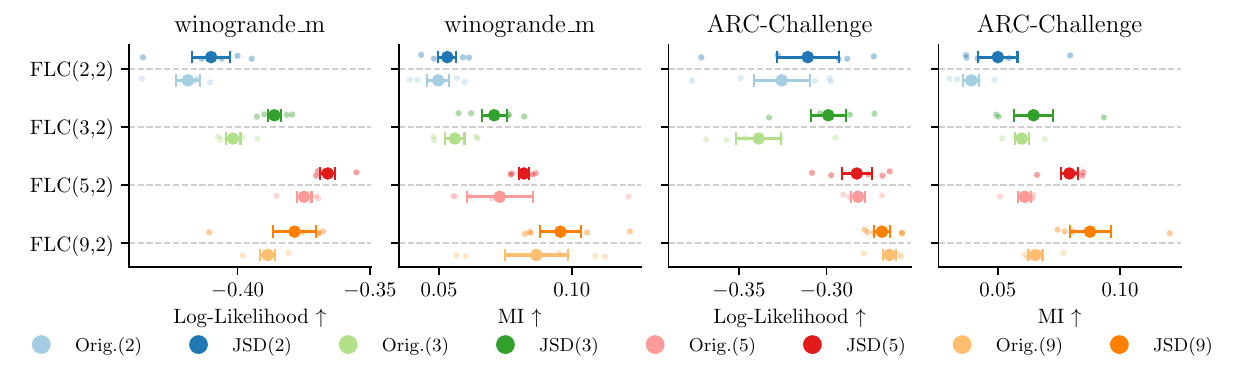}
    \caption{Functional diversity via JSD regularization. Comparison of Log-Likelihood ($\uparrow$) and mutual information (MI) ($\uparrow$) across two datasets (panels) for varying FLC model complexities (y-axis). Colors distinguish between the unregularized baseline (Orig., lighter shades) and the JSD-regularized models (JSD, darker shades). Solid points with error bars indicate the mean and standard error, while faded dots represent individual runs. The addition of the repulsive JSD regularizer significantly improves MI while maintaining comparable or slightly improved Log-Likelihood.}
    \label{fig:app_performance_jsd}
\end{figure}

\clearpage

\subsection{Additional smoothness on OOD examples}

\begin{figure}[ht]
    \centering
    \includegraphics[width=1.\linewidth]{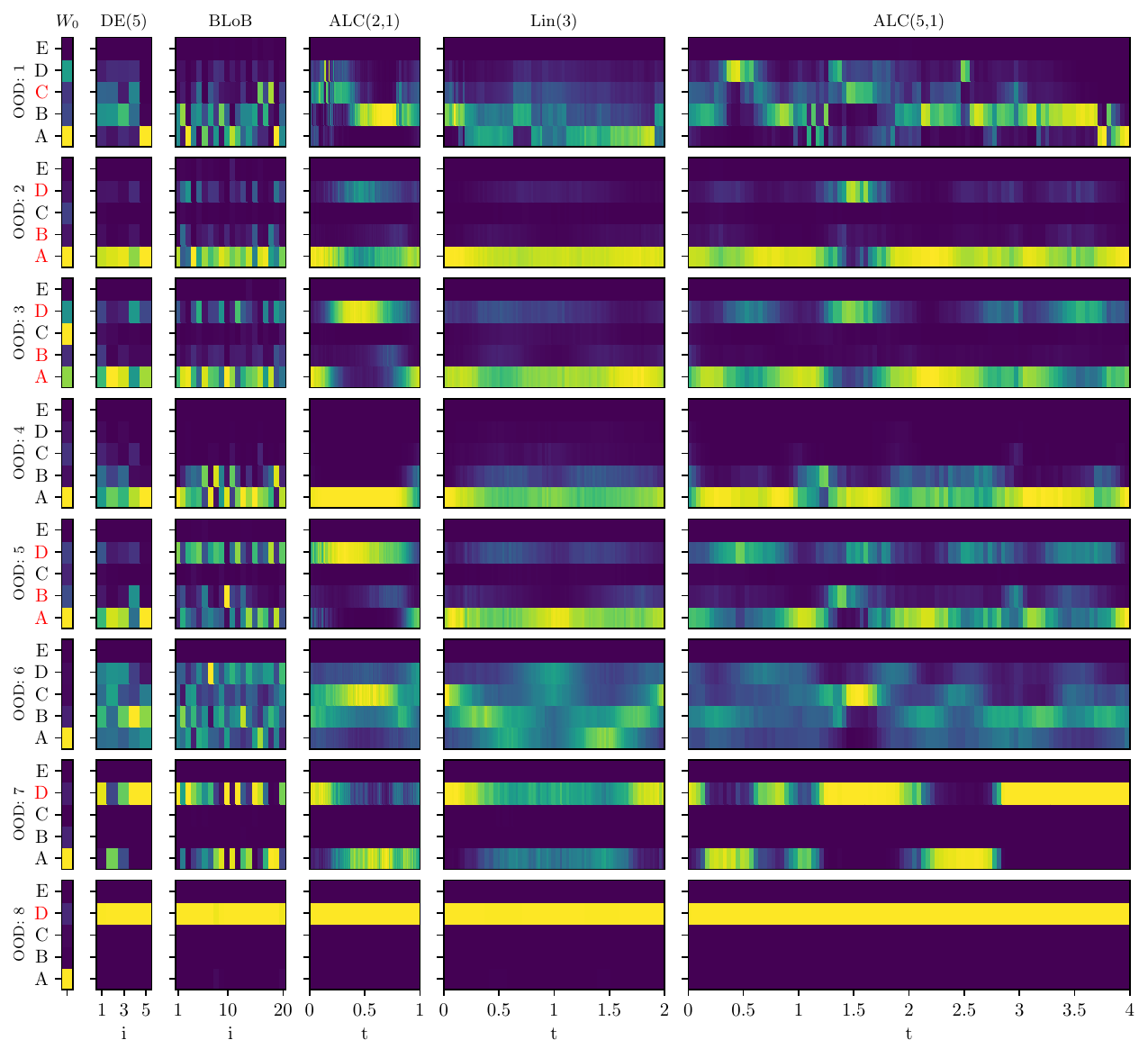}
    \caption{The class probability evolution for OOD examples using our fine-tuned LoRA model on the ARC-Challenge data. Predicted probabilities (linear color-coded from dark blue for $0$ to yellow for $1$) across five class labels (A--E) for eight OOD test examples (rows). The red class on the y-axis depicts the correct ground-truth label for each respective example. Facet columns compare various approaches: the base Qwen 2.5 model without pretraining ($W_0$), a 5-member Deep Ensemble (DE(5)), the BLoB variational competitor, linear interpolation between 3 anchors (Lin(3)), and our continuous Anchored LoRA-Curve methods (ALC(2,1) and ALC(5,1)). The x-axis represents either the discrete model index ($i$) or the continuous curve parameter ($t$).}
    \label{fig:app_smootheness_OOD}
\end{figure}

To further rigorously evaluate predictive smoothness on highly uncertain data, we constructed specialized out-of-distribution (OOD) examples in which exactly one, multiple, or no candidate classes are correct (cf.\ \cref{app:ood_examples}). As shown in \cref{fig:app_smootheness_OOD}, we consistently observe smooth transitions between class probabilities despite these distributional shifts. Notably, even when no valid class is provided, the predictive distribution can still transition smoothly across all candidate labels (e.g., OOD Example 6). However, we note that all evaluated methods occasionally exhibit failure modes on these extreme edge cases, as demonstrated by the breakdown in OOD Example 4 (cf.\ \cref{fig:app_smootheness_OOD}).

\subsubsection{OOD examples}

To rigorously evaluate predictive behavior under distributional shift, we generated a custom set of out-of-distribution (OOD) examples structurally aligned with the ARC-Challenge dataset. Using Gemini, we carefully crafted questions that deliberately violate the dataset's standard single-answer assumption by requiring multiple correct labels (e.g., assigning ground-truth classes \texttt{[0, 1, 3]}). This multi-label extension provides a realistic simulation of OOD scenarios, forcing the model to navigate edge cases where standard predictive confidence typically breaks down.

\label{app:ood_examples}
\begin{verbatim}

  - id: 1
    prompt:
      Return the label of the correct answer for the question below.

      Question: What is the value of x in the equation 3x = 12?
      Choices:
      2) A
      3) B
      4) C
      5) D
      Answer:
    label: 2

  - id: 2
    prompt:
      Return the label of the correct answer for the question below.

      Question: Which of the following are renewable energy sources?
      Choices:
      solar power) A
      wind power) B
      coal) C
      hydroelectric power) D
      Answer:
    label: [0, 1, 3]

  - id: 3
    prompt:
      Return the label of the correct answer for the question below.

      Question: Which of the following statements correctly describe properties 
      of democratic systems?
      Choices:
      citizens can vote in elections) A
      laws apply equally to all citizens) B
      power is inherited through family lineage) C
      governments are accountable to the public) D
      Answer:
    label: [0, 1, 3]

  - id: 4
    prompt:
      Return the label of the correct answer for the question below.

      Question: What caused the event described above?
      Choices:
      friction) A
      gravity) B
      magnetism) C
      electricity) D
      Answer:
    label: []

  - id: 5
    prompt:
      Return the label of the correct answer for the question below.

      Question: Which of the following are mammals?
      Choices:
      whale) A
      bat) B
      lizard) C
      human) D
      Answer:
    label: [0, 1, 3]

  - id: 6
    prompt:
      Return the label of the correct answer for the question below.

      Question: Which answer choice is correct for this question?
      Choices:
      this one) A
      this one) B
      this one) C
      this one) D
      Answer:
    label: []

  - id: 7
    prompt:
      Return the label of the correct answer for the question below.

      Question: A square has four equal sides and no equal sides. What shape is it?
      Choices:
      square) A
      circle) B
      triangle) C
      none of the above) D
      Answer:
    label: 3

  - id: 8
    prompt:
      Return the label of the correct answer for the question below.

      Question: Which of the following has the highest moral responsibility?
      Choices:
      a rock) A
      a tree) B
      a river) C
      a conscious human) D
      Answer:
    label: 3

\end{verbatim}

\clearpage


\section{Discarded approach}

\subsection{Repulsive regularization}
\label{app:repulsive-regularization}

Early in the development of the ALC framework, we hypothesized that enforcing geometric diversity directly in the parameter space would prevent mode collapse. To this end, we introduced a repulsive regularizer that penalized the pairwise cosine similarity between the curve's control points, $\vartheta_i$. For $N$ control points, this regularizer was defined as the sum of squared cosine similarities over all distinct pairs:
\begin{equation*}
    \mathcal{R}_{\text{repulsive}} = \sum_{1 \le i < j \le N_{\mathrm{cp}}} \left( \frac{\vartheta_i \cdot \vartheta_j}{\|\vartheta_i\| \|\vartheta_j\|} \right)^2.
\end{equation*}
While this explicitly forced the control points to disperse across the LoRA space, it yielded inconsistent downstream performance. Crucially, geometric distance in weight space did not strongly correlate with functional diversity in the model's predictive distributions. Instead, strong repulsive forces often pushed the control points into higher-loss regions, causing the trajectory to deviate from the optimal low-loss valley. 

Consequently, this weight-space regularization was discarded in favor of output-space regularization (such as the JSD regularizer described in \cref{subsec:jsd_math}), which directly encourages diversity in the predictive distributions without unnecessarily distorting the underlying geometric path.

\subsection{JSD regularization with corrupted input}
Further, we explored a variant of the JSD regularizer that maximizes functional diversity by explicitly enforcing predictive disagreement on out-of-distribution (OOD) inputs. Rather than evaluating the regularizer on clean training data, we computed the Jensen-Shannon divergence on dynamically corrupted inputs $\tilde{x}$. We evaluated two corruption strategies: replacing a random subset of input tokens with uniform random tokens from the vocabulary, and dynamically dropping a fraction of attention-masked tokens. For both variants, the loss was computed as:
$$
\mathcal{L} = \frac{1}{2}\big(\mathcal{L}_{\mathrm{CE}}(t_1) + \mathcal{L}_{\mathrm{CE}}(t_2)\big) + \lambda_{\mathrm{JSD}} \max(\tau_{\mathrm{JSD}} - D_{\mathrm{JSD}}(\tilde{p_{t_1}} \parallel \tilde{p_{t_2}}), 0),
$$
where $t_2 = (t + t_{\max}/2) \pmod{t_{\max}}$ ensures maximal separation along the curve, and $\tilde{p_t} := \operatorname{softmax}(f(\tilde{x}; W_{\vartheta}(t)))$ denotes the predictive distribution on the corrupted input. 

The intuition was that forcing disagreement on OOD inputs would encourage the curve to explore a broader range of functional regions. Empirically, this formulation successfully achieved its direct objective: the JSD measured on the corrupted inputs $\tilde{x}$ increased substantially during training. However, this synthetic functional diversity did not translate to significant improvements in generalization performance. Because the severely distorted inputs $\tilde{x}$ lay too far outside the true data manifold.

Furthermore, this approach introduced a severe computational bottleneck. Evaluating the OOD disagreement alongside the standard loss necessitated four forward passes per gradient update step. This pushed memory requirements to the limit, restricting the feasibility of training exclusively on high-capacity hardware (e.g., H100 GPUs with 94GB VRAM). 

Consequently, we reverted to the clean-data JSD regularizer (Section~\ref{subsec:jsd_math}), which directly enforces diversity on the task-relevant input distribution without the prohibitive variance or excessive memory overhead of synthetic OOD corruption. Nevertheless, explicit OOD disagreement remains a promising future direction for architectures fundamentally designed to leverage heavily augmented views, such as contrastive learning frameworks.

\section{Hyperparameter selection}
\label{sec:app_hparam-selection}

We largely align our core hyperparameters with those established by ScalaBL \citep{samplawskiScalable2025}. However, to ensure reliable convergence of the curve, we extend the training budget to 10,000 update steps, utilizing a one-cycle learning rate schedule peaking at $10^{-4}$ alongside early stopping on a 10\% validation split. Crucially, this extended and regularized optimization regime explains why our MAP and DE baselines achieve significantly stronger performance than those previously reported by ScalaBL. 

Furthermore, to accurately approximate the continuous trajectory, we dynamically scale the number of equidistant grid points $M$ based on the curve's complexity by setting $M = 2N_{\mathrm{cp}} - 1$. This formulation guarantees sufficient evaluation density, even for highly segmented curves, by explicitly placing one grid point at every control point---capturing its peak influence on the path---and one grid point midway between each adjacent pair.

All experiments were logged with Weights \& Biases (W\&B). Hyperparameters were explored using W\&B sweeps. 
The hyperparameter search space and fixed settings are detailed in the following:

\begin{table*}[htbp]
\centering
\caption{Hyperparameters varied in the W\&B project and their methodological role.}
\label{tab:varied-hparams}
\begin{tabular}{p{0.23\textwidth} p{0.18\textwidth} p{0.5\textwidth}}
\toprule
\textbf{Hyperparameter} & \textbf{Observed Values} & \textbf{Role in Method} \\
\midrule
\texttt{num\_curve\_segment} & \texttt{1, 2, 4, 8} & Controls the number of segments $\nseg$. Higher values allow the curve to navigate more complex landscapes. \\
\texttt{SegDeg} & \textbf{1, 2, 3,} \texttt{4, 6, 8, 12, 16, 24} & Degree of a single B\'ezier segment (similar to $m+1$). \\
\texttt{Pretraining} & [True, False] & If True it uses ALC otherwise FLC \\
\texttt{rng\_seed} & \texttt{1, 2, 20, 400, 8000} & Random seeds for initialization and sampling. \\
\bottomrule
\end{tabular}
\end{table*}

\begin{table*}[htbp]
\centering
\caption{Fixed hyperparameters and configurations maintained across all experiments. JSD hyperparameters only for JSD experiments. These parameters were also optimized through a previous small grid analysis on the \textit{Winogrande s} validation dataset.}
\label{tab:fixed-hparams}
\begin{tabular}{p{0.13\textwidth} p{0.19\textwidth} p{0.15\textwidth} p{0.40\textwidth}}
\toprule
\textbf{Category} & \textbf{Hyperparameter} & \textbf{Value} & \textbf{Description} \\
\midrule
\multirow{5}{*}{LoRA} 
    & $r$ & $8$ & Dimensionality of the low-rank updates. \\
    & $\alpha$ & $16$ & Alpha scaling factor applied to the LoRA matrices. \\
    & \texttt{lora\_layers} & value, query, lm\_head & Layer names of Qwen2.5 which are adapted. \\ 
\midrule
\multirow{5}{*}{Optimization} 
    & \texttt{optimizer} & AdamW & Algorithm utilized for gradient descent. \\
    & \texttt{weight\_decay} & $0.001$ & Regularization penalty applied to weights. \\
    & \texttt{val\_percentage} & $10$\% \newline(BoolQ 2.5\%) & Percentage of validation date out of original train data. \\
\midrule
\multirow{6}{*}{LR Schedule} 
    & \texttt{type} & One-cycle & Learning rate schedule policy. \\
    & \texttt{steps} & $10,000$ & Total duration of the training schedule. \\
    & \texttt{peak\_value} & $0.0001$ & peak learning rate. \\
    & \texttt{pct\_start} & $0.12$ & Percentage of steps spent increasing the learning rate. \\
    & \texttt{div\_factor} & $300$ & Initial learning rate division factor. \\
\midrule
Training & \texttt{batch\_size} & $4$ & Number of samples processed per training step. \\
\midrule
\multirow{2}{*}{Flat-LoRA} 
    & $\rho$ & $0.25$ & Magnitude of the perturbation noise. \\
    & \texttt{rho\_sched\_freq} & $50$ & Frequency for noise scheduling. \\
\midrule
\multirow{2}{*}{JSD} 
    & $\lambda_{\mathrm{JSD}}$ & $0.2$ & Magnitude of the JSD regularization. \\
    & $\tau_{\mathrm{JSD}}$ & $0.05$ & Cut-off value for JSD \\
\bottomrule
\end{tabular}
\end{table*}

\section{Computational environment and performance metrics}
\label{sec:appendix_environment}

This section details the hardware, software, and runtime performance characteristics of our LoRA-Curve experiments.

\subsection{Hardware and software specifications}

All experiments were conducted on high-performance compute nodes equipped with the following NVIDIA A100 (40GB VRAM) and NVIDIA L40S (48GB VRAM) GPU nodes.

The software stack utilized for the implementation and orchestration of experiments included:
\begin{itemize}
    \item \textbf{Language:} Python 3.10.18.
    \item \textbf{Frameworks:} JAX 0.5.0, Flax 0.10.0, and PyTorch (for weight loading and data processing).
    \item \textbf{Tracking:} Weights \& Biases (W\&B) for experiment tracking and metric logging.
\end{itemize}

\subsection{Model and training statistics}

The base model used for all experiments is Qwen2.5 7B \citep{yang2024qwen2}, which contains approximately 7.65 billion parameters. When fine-tuned using LoRA parameters in float32 precision and original Qwen2.5 $W_0$ in bfloat16 precision, the total parameter memory footprint is approximately 14.6 GB. The number of trainable parameters varies with the curve resolution:
\begin{itemize}
    \item \textbf{Anchored LoRA-Curve (ALC):} Since the anchors are frozen, only the intermediate handle points are trainable. For a rank $r=8$ configuration, trainable parameters range from 11M to 64M depending on the number of anchors $N$ and handles per segment $m$.
    \item \textbf{Training Budget:} Full curves (FLC and ALC) are optimized for 10,000 steps. In ALC, the individual anchor points (corresponding to DE members) are independently pretrained for 5,000 steps each.
\end{itemize}

\subsection{Runtime comparison}

Across our W\&B project, we logged over 980 runs, for which we report typical training times across different curve configurations in \cref{tab:runtimes}. The total compute investment for the results presented in this work is estimated at approximately 2500 GPU-hours. 

Note that the FLC walltime decreases for simpler curves because we dynamically scale the number of equidistant grid points in our approximation based on the number of control points. Furthermore, while FLC theoretically avoids the $N\times$ sequential training overhead inherent to pretraining ALC anchors, the empirical speedup does not scale exactly by a factor of $N$. This discrepancy arises because the reported wall times include fixed JAX compilation and evaluation overheads, and because individual ALC anchors are only pretrained for 5,000 steps, whereas the curve optimization runs for the full 10,000 steps.

\begin{table}[ht]
    \centering
    \caption{Runtime for a single experiment for different LoRA-Curve configurations and datasets. Times are reported for 10,000 FLC training steps on a single NVIDIA L40S GPU. The reported time includes training time, evaluation of test performance, and an additional evaluation using the training and validation data, which are only used for diagnostic purposes.}
    \label{tab:runtimes}
    \begin{tabular}{lcc}
        \toprule
        \textbf{Method}$(N, m)$ & Dataset &  \textbf{Time (h)} \\
        \midrule
        $\text{ALC}(9,2)$ & BoolQ & 18h 50m  \\
        $\text{FLC}(9,2)$ & BoolQ & 12h 10m  \\
        $\text{ALC}(2,2)$ & BoolQ & 6h 10m  \\
        \midrule
        $\text{ALC}(9,2)$ & OBQA & 5h 50m  \\
        $\text{FLC}(9,2)$ & OBQA & 3h 15m  \\
        $\text{ALC}(5,2)$ & OBQA & 4h 50m  \\
        $\text{FLC}(5,2)$ & OBQA & 2h 44m  \\
        $\text{ALC}(3,2)$ & OBQA & 3h 23m  \\
        $\text{FLC}(3,2)$ & OBQA & 2h 30m  \\
        $\text{ALC}(2,2)$ & OBQA & 2h 55m  \\
        $\text{FLC}(2,2)$ & OBQA & 2h 20m  \\
        \midrule
        $\text{ALC}(9,2)$ & Winogrande m & 3h 40m \\
        $\text{FLC}(9,2)$ & Winogrande m & 1h 35m \\
        $\text{ALC}(2,2)$ & Winogrande m & 1h 39m \\
        $\text{FLC}(2,2)$ & Winogrande m & 1h 10m \\
        \bottomrule
    \end{tabular}
\end{table}

\end{document}